\documentclass[journal]{IEEEtran}
\usepackage{times}
\usepackage{epsfig}
\usepackage{graphicx}
\usepackage{amsmath}
\usepackage{amssymb}
\usepackage{tabularx}
\usepackage{multirow}
\usepackage{color}

\usepackage{cite}
\usepackage{amsmath}
\usepackage{color}

\begin{document}

\title{Dual Convolutional LSTM Network for Referring Image Segmentation}

\author{Linwei Ye,~
        Zhi Liu,~\IEEEmembership{Senior Member,~IEEE,}
        and~Yang Wang

\thanks{
\textsuperscript{\textcopyright} 2020 IEEE.  Personal use of this material is permitted.  Permission from IEEE must be obtained for all other uses, in any current or future media, including reprinting/republishing this material for advertising or promotional purposes, creating new collective works, for resale or redistribution to servers or lists, or reuse of any copyrighted component of this work in other works.

L. Ye and Y. Wang are with the Department of Computer Science, University of Manitoba, Winnipeg, MB R3T 2N2, Canada (e-mail: \{yel3, ywang\}@cs.umanitoba.ca).

Z. Liu is with Shanghai Institute for Advanced Communication and Data Science, and School of Communication and Information Engineering, Shanghai University, Shanghai 200444, China (e-mail: liuzhisjtu@163.com).}}

\markboth{Journal of \LaTeX\ Class Files,~Vol.~, No.~, ~2020}%
{Shell \MakeLowercase{\textit{et al.}}: Bare Demo of IEEEtran.cls for IEEE Journals}

\maketitle

\begin{abstract}
We consider referring image segmentation. It is a problem at the intersection of computer vision and natural language understanding. Given an input image and a referring expression in the form of a natural language sentence, the goal is to segment the object of interest in the image referred by the linguistic query. To this end, we propose a dual convolutional LSTM (ConvLSTM) network to tackle this problem. Our model consists of an encoder network and a decoder network, where ConvLSTM is used in both encoder and decoder networks to capture spatial and sequential information. The encoder network extracts visual and linguistic features for each word in the expression sentence, and adopts an attention mechanism to focus on words that are more informative in the multimodal interaction. The decoder network integrates the features generated by the encoder network at multiple levels as its input and produces the final precise segmentation mask. Experimental results on four challenging datasets demonstrate that the proposed network achieves superior segmentation performance compared with other state-of-the-art methods.

\end{abstract}


%
\IEEEpeerreviewmaketitle

\section{Introduction}
\label{sec:intro}
\IEEEPARstart{S}{egmenting} objects of interest in an image is a fundamental problem in computer vision field. Researchers have formulated various high-level computer vision tasks related to segmenting objects in images, such as semantic segmentation~\cite{long2015fully,chen2016deeplab}, instance segmentation~\cite{he2017mask}, salient object segmentation~\cite{cheng2014global,ye2017salient}. However, these tasks have some limitations. For example, semantic segmentation and instance segmentation assume a pre-defined set of object categories (e.g., cat, person, bus, etc). Salient object segmentation does not have the restriction on the pre-defined object categories and is based on the human visual cognition system to distinguish the most salient objects in the scene from background. But in some complicated cases, the salient objects are ambiguous and may not be unique for different viewers~\cite{li2014secrets}.

\begin{figure}[!ht]
  \centering
{\setlength{\tabcolsep}{0.5pt}
\scalebox{0.88}{
  \begin{tabular}{ccc}

    man with back to camera & guy on left & man in yellow shirt \\
    
    \includegraphics[width=1.2in]{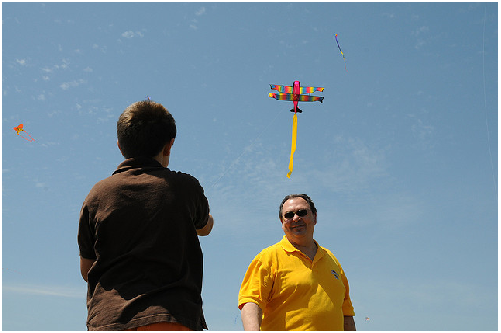}&
    \includegraphics[width=1.2in]{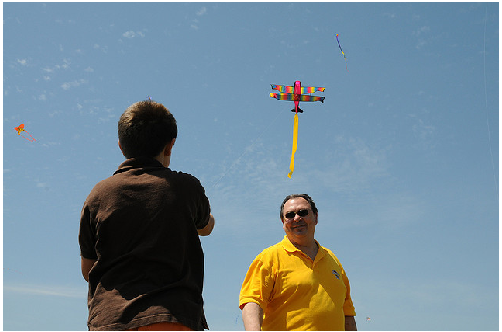}&
    \includegraphics[width=1.2in]{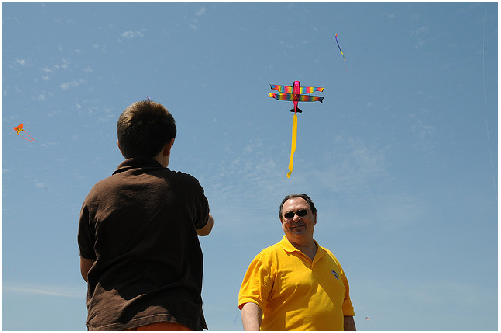}\\
    
    \includegraphics[width=1.2in]{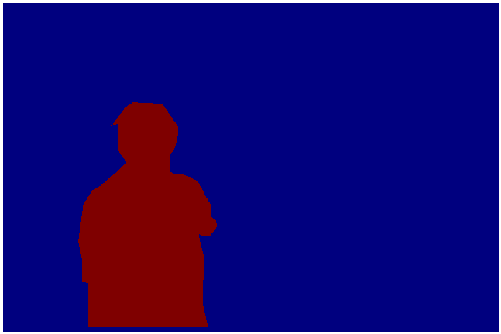}&
    \includegraphics[width=1.2in]{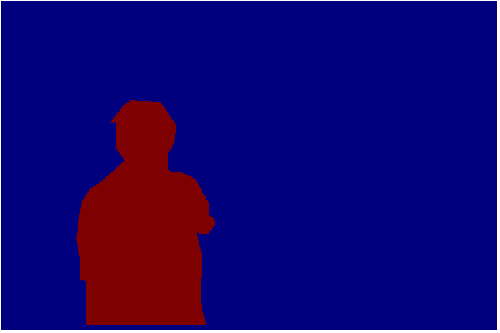}&
    \includegraphics[width=1.2in]{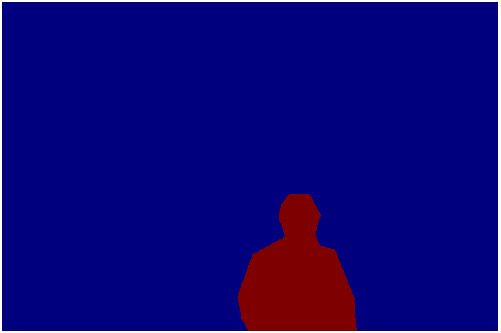}\\

  \end{tabular}
}
}
\caption{Illustration of the referring image segmentation task. Given an input image and a referring expression, the goal is to generate the segmentation mask for the referred object in the image. The referring expression may use diverse descriptions to identify the referred object, such as object names (e.g. ``man'', ``guy''), attributes (e.g. ``back to camera'', ``yellow'') and spatial relationships (e.g. ``on left''). The first row shows three query expressions and the last row indicates the corresponding segmentation mask in the image.}
\label{fig:intro}
\end{figure}

In recent years, referring image segmentation has attracted the attentions of many researchers. In referring image segmentation, the object of interest to be segmented is specified by a free-form referring expressions in natural language.  Fig.~\ref{fig:intro} illustrates the example of referring image segmentation. With an input image, the segmentation mask can be referred using diverse descriptions for the same object (the first two examples from the left) by its attribute ``back to camera'' or spatial relationship``on left'' to differ the two same categorical objects ``man'' or ``guy''. The prediction can also be identified by the attribute of the shirt ``yellow'' related to another object in the image. Referring image segmentation is a challenging problem which requires a combinational comprehension of both linguistic and visual information, and enables many real-world applications including interactive image editing, intelligence visual search and  human-robot interaction.

There are several existing works in this area. Some of them~\cite{hu2016segmentation,li2018referring,shi2018key} represent the whole referring expression and visual features separately. For example, the referring expression is encoded as a hidden vector using recurrent neural network (RNN) or long short-term memory (LSTM) model~\cite{hochreiter1997long}, while the input image is represented using convolutional neural network (CNN) features. The textual feature vector is then combined with visual features at each spatial location followed by deconvolution~\cite{hu2016segmentation}, recurrent refinement~\cite{li2018referring} and key-word context~\cite{shi2018key}  
for producing the final segmentation mask. The limitation of these approaches is that visual and textual features are extracted in an independent way. 
It may not be able to capture the detailed multimodal information often useful for the referring image segmentation task.

A different line of previous work~\cite{liu2017recurrent,margffoy2018dynamic} processes each word in the referring expression in a sequential order. These methods can potentially capture the detailed information of words in the referring expression with visual context by a sequential interaction~\cite{liu2017recurrent} or synthesis module~\cite{margffoy2018dynamic}. However, these methods treat every word equally in their models. This may cause difficulty for long referring expressions that likely contain unimportant words.

To address the limitations of previous work, we propose a dual convolutional LSTM network for referring image segmentation. 
The convolutional LSTM (ConvLSTM) is originally proposed in~\cite{xingjian2015convolutional} to replace fully connected layers with convolutional layers in order to capture spatial-temporal information of a sequence of images. We elaborately modify the original ConvLSTM to fit the multimodal data in the referring segmentation problem by the proposed encoder-decoder framework. Our network models the input image as the spatial information, and formulates the language expression and multi-level features as the temporal sequence for the encoder and the decoder, respectively.
Specifically, the encoder network (E-ConvLSTM) is used to capture multimodal feature interactions. It first adopts the feature maps of the input image as the spatial information. For each spatial location of the feature map, we capture the interaction at this location over every word in the referring expression at each recurrent time step of E-ConvLSTM to gradually localize the referred object. In addition, the proposed approach introduces word-level attentions embedded into the cell state of E-ConvLSTM to guide the interaction process towards more important words (e.g. words corresponding to the object of interest), instead of treating each word equally in the recurrent step~\cite{liu2017recurrent,margffoy2018dynamic}.
The decoder network (D-ConvLSTM) formulates attentive multimodal features encoded by E-ConvLSTM at different levels as the sequence and iteratively refines these features to take full advantage of correlations from multi-level features. We further introduce spatial attentions for multi-level features to better focus on the semantics by the high-level features and fine details by the low-level features for a precise segmentation mask.

In summary, the main contributions of this paper lie in the following four aspects:

1) We propose a dual convolutional LSTM network to exploit an encoder-decoder framework for multimodal feature encoder and multi-level segment decoder in the spatial regions for referring image segmentation.

2) The multimodal feature encoder embeds word attention into the cell (memory) state of E-ConvLSTM to adaptively encode the multimodal interaction towards more important words and localize the referred objects.

3) The multi-level segment decoder (D-ConvLSTM) progressively decodes the features with spatial attention at multiple levels to refine more precise referring image segmentation results.

4) Our proposed dual convolutional LSTM network is evaluated on four public available datasets thoroughly and achieves the state-of-the-art performance.

The rest of this paper is organized as follows. We first introduce some related works in Sec.~\ref{sec:related} and the basic ConvLSTM unit as background in Sec.~\ref{sec:clstm}. Then we present an encoder-decoder framework consisting of the proposed E-ConvLSTM in Sec.~\ref{sec:lclstm} and D-ConvLSTM in Sec.~\ref{sec:vclstm} in detail. Experimental setup and extensive experimental results for performance evaluation are given in Sec.~\ref{sec:experiment} and Sec.~\ref{sec:result}, respectively. Finally, we make conclusions in Sec.~\ref{sec:conclude}.

\section{Related Work}\label{sec:related}
In this section, we first review some relevant object segmentation tasks. Then a set of vision and language problems  is introduced for studies in the intersection of multimodal information processing. Last, the recent studies tightly related to our work about referring image segmentation are presented.

\noindent{\bf Object Segmentation:}  
Object segmentation or extraction from an image can be done rapidly and freely by human but is a challenge problem in computer vision. A major difficulty lies in that computers have to be aware of the object of interest ahead of executing segmentation operation. One straightforward way is to roughly specify an object interactively by human, e.g., drawing a bounding-box~\cite{rother2004grabcut}. Then a segmentation approach can refine the specific object with well-defined boundaries according to visual cues of the candidate region and background (inside and outside the bounding-box). The desired object can also be identified automatically inspired by human visual system which distinguish visually salient objects from background~\cite{liu2012unsupervised,li2014secrets,cheng2014global,ye2017salient}. In the recent data exploding era,  convolutional neural network~\cite{krizhevsky2012imagenet} drives significant segmentation performance boost in semantic segmentation~\cite{long2015fully,chen2016deeplab} where all pixels are labeled with pre-defined object categories, and instance segmentation~\cite{he2017mask} where additional instance labels are available. This makes visual understanding foundation for referring image segmentation in this paper.

\noindent{\bf Combining Vision with Language:} There has been a lot of previous works on combining vision with language for different tasks, such as image captioning~\cite{xu2015show} and visual question answering (vqa)~\cite{antol2015vqa}. These models incorporate attention mechanism to explore relevant visual features on the corresponding spatial regions, the localized regions are not precise enough since the goal of these works is to generate a sentence or a bounding box. 
Visual grounding~\cite{rohrbach2016grounding,yu2018mattnet} requires an exact bounding-box as the output according to a referring expression. A joint multimodal embedding is used in reconstruction of the text phrase~\cite{rohrbach2016grounding} and modular networks of subject, location and relationship~\cite{yu2018mattnet}. However, all these grounding methods rely on proposals generated by off-the-shelf object detectors.

For tasks like referring image segmentation, we need to effectively represent the multimodal interaction between the linguistic and visual information and keep detailed spatial information in order to generate a segmentation mask. 
A straightforward way to combine visual and linguistic features is to use simple concatenation as~\cite{hu2016segmentation,li2018referring}, key-word context \cite{shi2018key} or cross-modal self-attention mechanism \cite{ye2019cross} for multimodal features. 
Another line of work~\cite{liu2017recurrent,margffoy2018dynamic}, which is more related to our work, exploits the sequential nature of language and differ from how human solves this problem~\cite{underwood2004inspecting}. However, these two methods consider each word equally in the interaction. Instead, we exploit word attentions over the input expression and embed them into cell states of ConvLSTM to guide the effective multimodal feature interaction for the encoder network.

\noindent{\bf Referring Image Segmentation:} Referring image segmentation is first introduced by~\cite{hu2016segmentation} to segment the object-of-interest referred by an expression. They concatenate visual, linguistic and spatial features on the spatial feature maps and use a deconvolutional layer to recover a high-resolution segmentation mask. 
In \cite{liu2017recurrent}, a recurrent multimodal interaction model is proposed to gradually ground the referred objects onto the image according to the progressive meaning of the language input. Each word is gradually combined with  visual features in a sequential order. 
The work in \cite{margffoy2018dynamic} adopts a similar method as~\cite{liu2017recurrent} but generates dynamic filters in the synthesis module to incorporate linguistic information. It further uses an incremental module with bilinear upsampling and convolution over feature maps for fine details. 
However, these works equally encode every word in the referring expression at each step of a recurrent model. It may be difficult to effectively capture the important words in a long referring sentence. In this paper, we propose a multimodal feature encoder to embed word attentions into the cell state of ConvLSTM. The multimodal interaction of visual and linguistic features can be learned adaptively towards more important words in the expression to identify the referred object.

Key-word-aware context is proposed in~\cite{shi2018key} to combine textual feature with image regions to model their relationships, which aligns key words to different image regions. In order to obtain a more precise mask for segmentation,
more elaborated refinement approaches are proposed by~\cite{li2018referring} and~\cite{ye2019cross}. Specifically, multi-scale features are progressively refined to improve the segmentation mask from a roughly localized mask in~\cite{li2018referring}. In~\cite{ye2019cross}, the cross-modal self-attention network is proposed to capture the long-range dependencies between linguistic and visual contexts and a gated multi-level fusion is then used to extract a precise segmentation mask. 
Both methods show the effectiveness of the iterative refinement for removing irrelevant regions and producing more precise segmentation masks in the end. Though significant improvements have been achieved by these refinement methods, the importance and relation between different levels of features are not fully exploited.
Different from these methods which simply integrate multi-level visual features for refinement, we deploy a decoder to utilize encoded multimodal information and introduce spatial attentions into multi-level features to focus on more specific features in the refinement. It selectively concentrates on the main body of the referred object and boundary details according to the high-level and low-level features, respectively.

\noindent{\bf  Long Short-Term Memory Network:}
Long short-term memory (LSTM) network~\cite{hochreiter1997long} has been widely adopted for sequential data (e.g., language~\cite{sutskever2014sequence}, audio~\cite{marchi2014multi} and video~\cite{srivastava2015unsupervised}). LSTM can effectively capture long-range dependencies in sequential data. LSTM contains fully connected layers in both the input-to-state and state-to-state transitions with four gates including an input gate, a memory gate, a forget gate and an output gate. It can be flexibly used in both the encoder and the decoder framework for sequential inputs and outputs~\cite{sutskever2014sequence}. LSTM can be used to combine word features with visual context for multimodal interaction~\cite{xu2015show,liu2017recurrent}. Instead of propagating information in one direction as in standard LSTM, bidirectional LSTM is proposed to learn hidden states from both forward and backward directions at the same time. The forward and backward passes can learn comprehensive information through the sequential input~\cite{graves2005framewise,sundermeyer2014translation}. Convolutional LSTM (ConvLSTM) replaces the fully connected layers of LSTM with convolutional layers to make the sequential learning possible in spatial-temporal domains. It performs better in handling spatio-temporal correlations for a set of images~\cite{xingjian2015convolutional}. In addition, several other variants of LSTM are designed to stack multilayer LSTM for skeleton-based action recognition~\cite{zhang2017geometric}, formulate AutoEncoder topology~\cite{gensler2016deep} or combine fully convolutional neural networks with LSTM for vehicle counting~\cite{zhang2017fcn}.

\section{Background: Convolutional LSTM}\label{sec:clstm}
A  general model for referring image segmentation requires spatial relationships of an image and sequential dependency of words. Convolutional networks and LSTM \cite{hochreiter1997long} are two powerful feature representation approaches for an image and words, respectively. 
A convolutional network extracts hierarchical spatial semantic features from an image. LSTM models long-range dependency for sequential words. Convolutional LSTM (ConvLSTM)~\cite{xingjian2015convolutional} is an extension of the vanilla LSTM for capturing spatio-temporal relationship of data. It replaces fully connected layers in LSTM with convolutional layers for input-to-state and state-to-state transitions so that spatial correlation can also be built within the sequential inference process. 
In order to simultaneously capture spatial and sequential information, we introduce a dual convolutional LSTM framework for multimodal feature encoder and multi-level segment decoder between an image and words. The proposed network captures the interaction between linguistic and visual context needed for localizing the referred object. It also maintains the spatial information needed for producing a precise segmentation mask.

Let $\{X_1, X_2,...,X_T\}$ denote a set of feature maps where $X_{t}$ is the input at time step $t$ in ConvLSTM. The complete ConvLSTM operation can be summarized as follows:
\begin{eqnarray}
  &&
\begin{pmatrix} i_t \\ f_{t}\\ o_{t}\\ g_{t} \end{pmatrix}
= 
\begin{pmatrix} \sigma \\ \sigma \\ \sigma \\ \tanh \end{pmatrix}
W
\begin{pmatrix} X_{t} \\ H_{t-1} \end{pmatrix} \\
&& C_{t} = i_{t}\odot g_{t} + f_{t}\odot C_{t-1} \\
&& H_{t} = o_{t}\odot \tanh (C_{t})
\label{eq:hid}
\end{eqnarray}
where $W$ represents the ConvLSTM parameters, $\sigma$ is the sigmoid function and $\odot$ is the element-wise product. Here we use $i_{t}$, $f_{t}$, $o_{t}$, $g_{t}$ to denote the input gate, forget gate, output gate and memory gate, respectively. $H_{t}$ is the hidden state and $C_{t}$ is the cell state at each time step $t$. At each time $t$, ConvLSTM takes the input $X_t$ and previous hidden state $H_{t-1}$ to generate $i_{t}$, $f_{t}$, $o_{t}$ and $g_{t}$ at current time.

The essential design of ConvLSTM is to use convolution layer for spatial correlations and uses recurrence over time for sequential dependency. At each time $t$, the input gate $i_{t}$ controls how much information from $X_t$ is exposed to the cell state. The forget gate $f_{t}$ controls how much information from the past should be forgotten. This results in an updated cell state $C_t$. The output gate $o_{t}$ is then used to propagate effective information to the hidden state $H_t$ at time $t$ as the output. Therefore, the cell state $C_t$ is considered as memory state to control the information flow update. In other words, it determines how much the current and previous states influence the current hidden state $H_{t}$.

\begin{figure}[htb]
  \centering
  \includegraphics[width=8.2cm]{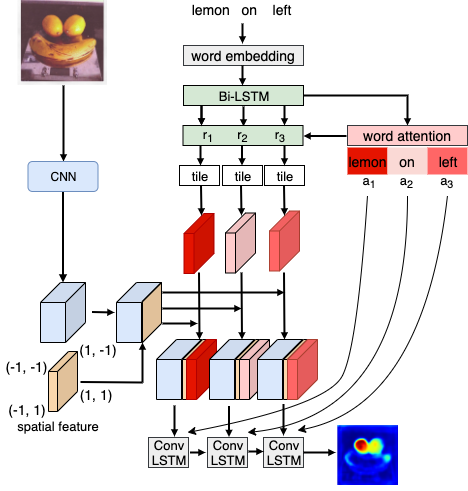}
\caption{The multimodal feature encoder with E-ConvLSTM. We embed word attentions into the ConvLSTM cell to adaptively encode the multimodal interaction towards more important words.}
\label{fig:lclstm}
\end{figure}

\begin{figure}[htb]
  \centering
  \includegraphics[width=8.8cm]{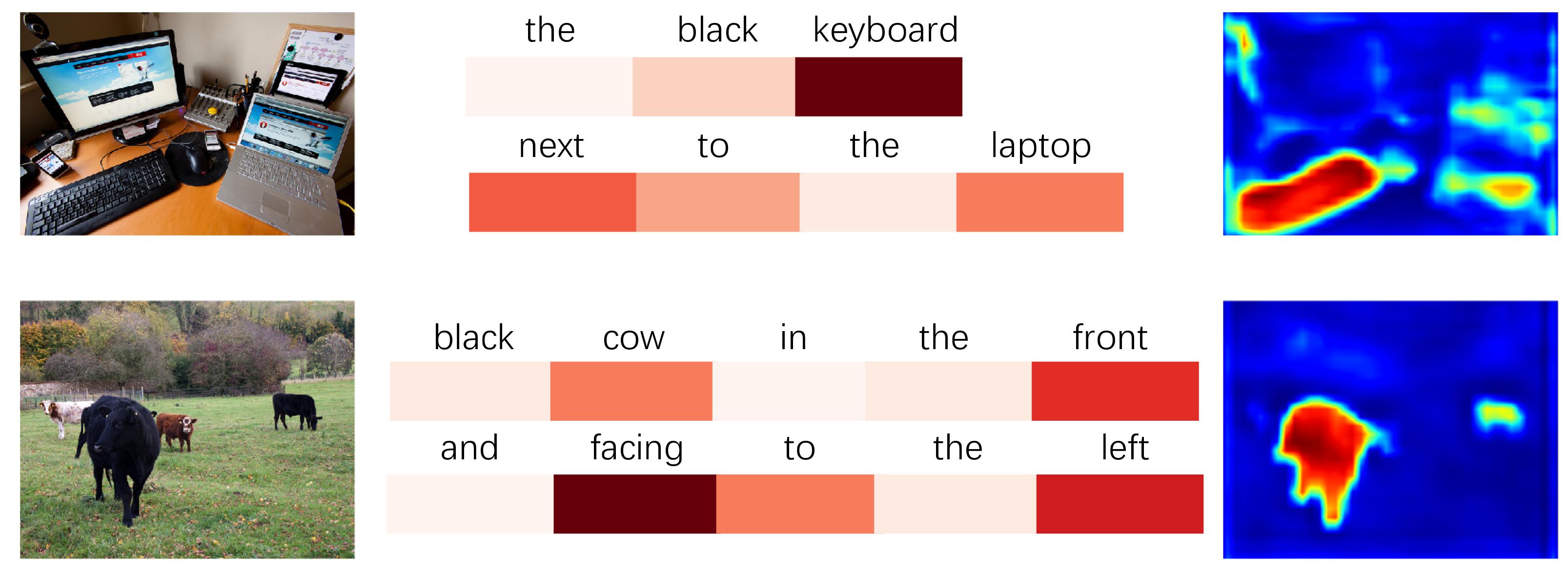}
\caption{Examples of word attentions. From left to right at each row: the original image, word attentions, visualization of the final feature map. Darker red color means higher attention weights.}
\label{fig:word_att}
\end{figure}

\section{Multimodal Feature Encoder}\label{sec:lclstm}
The goal of the encoder network of our model is to generate multimodal features that capture both detailed linguistic and visual information. The overall architecture of the encoder network is shown in Fig.~\ref{fig:lclstm}. First, we use an attention mechanism to learn to focus on certain important words in the referring expression (Sec.~\ref{sec:wa}). 
Then word vectors are re-weighed by word attentions and tiled as the same size as the visual feature map. Each word feature is concatenated with the visual feature  and the spatial feature for a word-specific multimodal feature. These multimodal features capture the complex interactions between the linguistic information from the referring expression and the visual information from the image. Finally, multimodal interaction (Sec.~\ref{sec:interact}) is guided by word attentions to adaptively towards more important words in the referring expression.

\subsection{Word Attention}\label{sec:wa}
Instead of representing the entire referring expression as a hidden vector~\cite{hu2016segmentation,li2018referring} using LSTM, we propose to keep track of the vector representation of each word in the expression. We also learn the relative importance (i.e. word attention) of each word and reweight the vector representation of each word with its corresponding attention score. This allows our model to focus on important words in the expression and use it to adaptively encode multimodal information.

Let us denote a referring expression as $\{w_1, w_2,..., w_L\}$ where $w_l$ is the $l$-th word and $L$ is the number of words. Each word $w_l$ is represented as a one-hot vector. We project each word into a vector representation $e_l$ using a word embedding layer, then use a bidirectional LSTM to produce a hidden vector for each word as follows: 
\begin{eqnarray}
&&e_l=embedding(w_l)\\
&&\overrightarrow{h}_{l} = \overrightarrow{LSTM}(e_l, \overrightarrow{h}_{l-1}) \\
&&\overleftarrow{h}_{l} = \overleftarrow{LSTM}(e_l, \overleftarrow{h}_{l+1})\\
&&h_l = [\overrightarrow{h}_{l}, \overleftarrow{h}_{l}]
\end{eqnarray}
where $\to$ and $\gets$ denote the forward and backward directions in bidirectional LSTM, respectively.  $h_l\in\mathbb{R}^{C_r}$ is used as the vector representation for the $l$-th word,  which concatenates the hidden states of bidirectional LSTM at every time step (word) $l$ with consideration of both previous and future relations of words.

We then apply two linear layers on the word feature $h_l$ and normalize the output to calculate an attention weight $a_l$ that indicates the relative importance of the word:
\begin{equation}
a_l = \frac{\exp(W_a^L \tanh( W_b^Lh_l))}{\sum_{k=1}^L\exp(W_a^L\tanh(W_b^Lh_k))}
\label{eq:al}
\end{equation}
where $W_a^L\in\mathbb{R}^{C_a}$ and $W_b^L\in\mathbb{R}^{C_a\times C_r}$ are the model parameters of these two linear layers. We then use the attention weight of each word to re-weight its vector representation as follows:
\begin{equation}
    r_l = a_l h_l
\label{eq:hl}
\end{equation}

The final vector representation $r_l$ is considered as the generated attentive word feature that takes into account of the relative importance of the $l$-th word. It conveys discriminative information in the given referring expression. Fig.~\ref{fig:word_att} shows examples of word attentions. The word attentions encourage the network to focus on the more important words so as to identify the referred object (``keyboard" in the top example) or discriminate the similar objects (``black cow" in the bottom example)  with the relative location (``front") and different attribute (``facing to the left''). These word attentions will also contribute differently on the multimodal interaction in the next section.

\subsection{Word Attentive Multimodal Interaction}\label{sec:interact}
We propose a word attentive multimodal interaction model that captures the multimodal information, including the linguistic information in the referring expression and the visual/spatial information in the input image.

\noindent\textbf{Multimodal Feature Generation:} We use a pretrained CNN network to extract a visual feature map for an input image. Let the feature map size be $W\times H\times C_v$, where $W$ and $H$ are the width and height of the feature map, and $C_v$ is the channel dimension. 
We then generate an $8$-dimensional vector representing the spatial information at each spatial location in the feature map as~\cite{liu2017recurrent}. 
Specifically, we use first three dimensions to encode the relative horizontal coordinates and another three dimensions to encode the relative vertical coordinates. The last two dimensions correspond to the relative sizes (width and height) of the image. The spatial feature map has a dimension of $W\times H\times 8$. All values of this spatial feature map are normalized to the range of $[-1, 1]$, i.e., the values of the upper left corner and the lower right corner of the spatial feature is $(-1, -1)$ and $(1, 1)$, respectively. 
This spatial feature is then appended to the visual feature map at each spatial location. Thus the dimension of the final visual feature map is $V\in\mathbb{R}^{W\times H\times (C_v+8)}$. This feature map captures both visual and spatial information of an image.

Then the multimodal feature can be generated as follows. For the $l$-th word, we append its vector representation $r_l$ to each spatial cell of the feature map $V$. This results in a word-specific multimodal feature $M_l^e\in\mathbb{R}^{W\times H\times (C_v+8+C_r)}$. We repeat this process for all words $\{r_1, r_2, ..., r_L\}$. In the end, we obtain $L$ word-specific multimodal feature maps ${\{M_1^e, M_2^e, ..., M_L^e\}}$.

\noindent\textbf{Multimodal Interaction:} Given the feature maps ${\{M_1^e, M_2^e, ..., M_L^e\}}$ obtained for each word separately, we want to combine these feature maps and capture their interactions along each word over time for detailed multimodal comprehension. Different from the previous work \cite{liu2017recurrent} that treats each word equally, our multimodal interaction model can pay more attention to important words by taking into account of word attentions.

For the word-specific multimodal feature maps $\{M_l^e\}^L_{l=1}$, we modify the standard ConvLSTM (see Sec.~\ref{sec:clstm}) in order to take into account of the word attentions as follows:
\begin{equation}\label{eq:lhidden}
(H_{l}^e, C_{l}^e) = ConvLSTM_E(M_l^e,H_{l-1}^e,C_{l-1}^e)  
\end{equation}
where
\begin{eqnarray}
C_{l}^e = a_l\times i_{l}^e\odot g_{l}^e + (1 - a_l)\times f_{l}^e\odot C_{l-1}^e
\label{eq:lcell}
\end{eqnarray}

The output of the hidden state $H_{l}^e$ at each time step summarizes the semantic information for all previous seen words before $l$. So the hidden state at the final step $H_{L}^e\in\mathbb{R}^{W\times H\times (C_s)}$ contains information of all words in the referring expression. The cell state $C_{l}^e$ can take advantage of the attention weight $a_l$ of the corresponding word. As shown in Eq.~\ref{eq:lcell}, the attention $a_l$ of a word (obtained in Eq.~\ref{eq:al}) is used to modulate the cell memory of the multimodal interaction. If a word has a higher attention weight, it will encourage the cell state to allow more information to flow from the current state. In contrast, a word with lower attention weight will allow less information flow into the cell state. So the cell state will rely more on the historic memory. In the end, the modified ConvLSTM is able to pay more attention to important words with higher attention weights. 

Fig.~\ref{fig:lclstm} illustrates the whole process of E-ConvLSTM. Given the input query ``lemon on left'', the multimodal feature encoder produces word attention $a_l$ for each word in the referring expression. The different word attentions are used to generate attentive word feature $r_l$ for multimodal features. In addition, the attention of  every word is also embedded into the cell memory of each E-ConvLSTM cell to adjust multimodal interaction towards more informative words (e.g., ``lemon'' and ``left'') adaptively.

\section{Multi-level Segment Decoder}\label{sec:vclstm}

Referring image segmentation aims to segment the referred object rather than predicting dense labels for the entire image. Directly applying the multimodal features to predict a segmentation mask might lead to an unsatisfactory result because of the distraction of non-referred regions. In addition, multi-level feature representations help to provide fine details for a more precise boundary of the object. 
To this end, we propose to generate spatial attention to focus on important spatial regions in the multimodal feature maps. 
The spatial attentions are then applied at the different levels of encoded multimodal features generated by the E-ConvLSTM. We also propose the multi-level segment decoder with another ConvLSTM (D-ConvLSTM) to sequentially refine the different levels of encoded multimodal features for a precise segmentation mask.

The hidden state $H_{L}^e$ of the last word from Eq.~\ref{eq:lhidden} can be used as the input to the decoder network. Note that $H_{L}^e$ is computed using the CNN features at a particular layer. In practice, we can apply the encoder by using the CNN features at several different layers in the network. We use $\{M_1^d, M_2^d, ..., M_S^d\}$ to denote the hidden state $H_L^e$ in the encoder based on different levels of CNN visual features. In this paper, we set $s=3$ and $\{M_1^d, M_2^d, M_3^d\}$ corresponding to \{$Res5, Res4, Res3$\} of a DeepLab101 backbone network~\cite{chen2016deeplab}. Specifically, $M_s^d\in\mathbb{R}^{W\times H\times (C_s)}$ can be generated by applying the encoder network in Sec.~\ref{sec:interact} at a particular level $s$ for multimodal representation over the referring expression. 

\begin{figure}[htb]
  \centering
{\setlength{\tabcolsep}{0.5pt}
\scalebox{0.85}{
  \begin{tabular}{cccccc}

    \multicolumn{4}{c}{Referring expression: \textit{``dude in blue shirt tie''}} \\
    \includegraphics[width=1.0in]{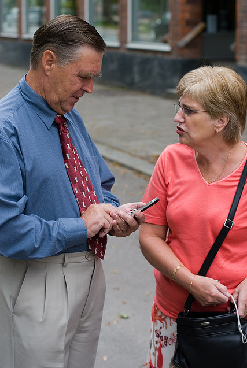}&
    \includegraphics[width=1.0in]{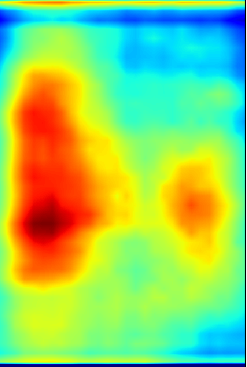}&
    \includegraphics[width=1.0in]{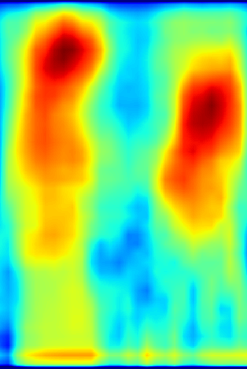}&
      \includegraphics[width=1.0in]{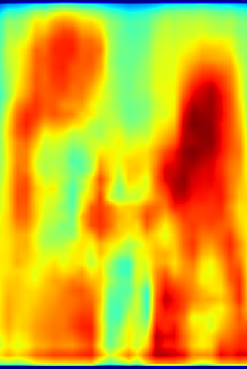}\\
      
       \multicolumn{4}{c}{Referring expression: \textit{``right meter''}} \\
    \includegraphics[width=1.0in]{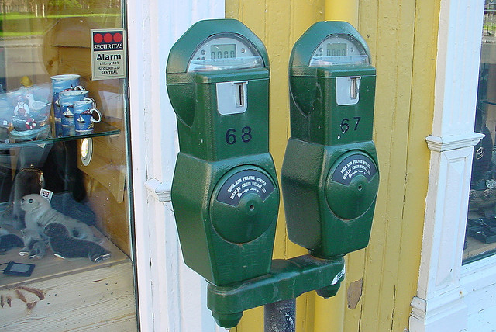}&
    \includegraphics[width=1.0in]{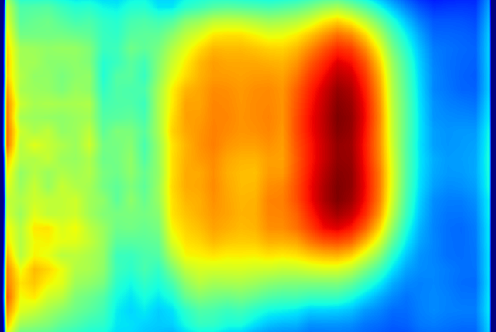}&
    \includegraphics[width=1.0in]{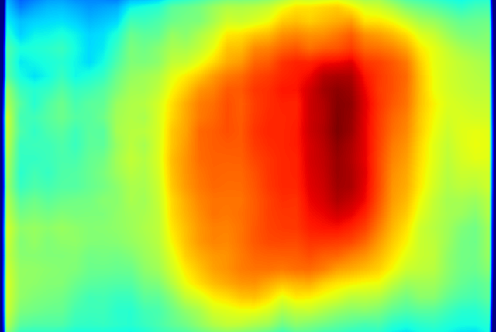}&
      \includegraphics[width=1.0in]{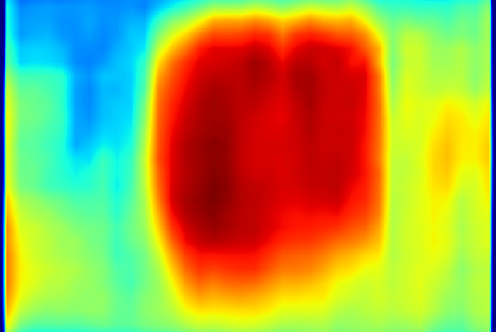}\\

      (a) & (b) & (c) & (d)\\
    
  \end{tabular}
}
}

\caption{Visualization of the spatial attentions corresponding to different levels of encoded features. (a) original image; (b,c,d) the spatial attentions from the high-level to low-level features.}
\label{fig:sp_att}
\end{figure}

\noindent\textbf{Spatial Attention:} The spatial attention is generated to focus on important spatial regions in the feature map. We use a convolutional layer with a large kernel ($7\times 7$) to capture relatively large regions. The convolution of a large kernel can effectively gather information with a larger receptive field and is robust for localization~\cite{peng2017large}.  
\begin{equation}
 M_s^d = \sigma (W_s\ast M_s^d+b_s)\odot M_s^d \\
\end{equation}
where $W_s$ and $b_s$ are the parameters of the convolution filters. The sigmoid function $\sigma$ summarizes  the importance of different regions in the feature map. Then it is applied to each slice of $M_s^d$. 
Fig.~\ref{fig:sp_att} presents the spatial attentions in different levels of features. It can be observed that spatial attentions can help the decoder refines features from the important spatial regions. The spatial attention corresponding to high-level features tends to concentrate on the referred objects, while the spatial attention of the low-level features tends to be more spread-out.

\begin{figure}[htb]
  \centering
  \includegraphics[width=8cm]{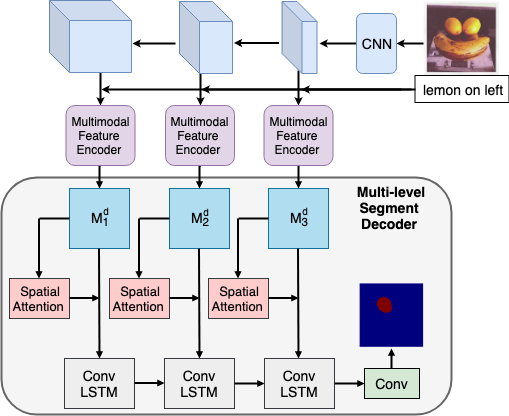}
\caption{An illustration of the multi-level segment decoder with D-ConvLSTM. The decoder network iteratively refines the segmentation mask by using the features extracted from different CNN levels. It also uses spatial attentions to focus on image regions that are informative for generating the segmention mask.}
\label{fig:vclstm}
\end{figure}

\noindent\textbf{Multimodal Feature Refinement:} Existing works~\cite{li2018referring,margffoy2018dynamic} have shown that refining the segmentation mask over multi-level features can significantly improve the performance with respect to fine boundaries and completed objects. Instead of directly taking visual features from different level layers of the network, we propose to use multimodal features $M_s^d$ that have incorporated word-attentive interactions and spatial attentions.

As shown in Fig.~\ref{fig:vclstm}, the decoder network progressively integrates these features $\{M_s^d\}_{s=1}^S$ from high-level to low-level semantics as follows:
\begin{equation}
(H_{s}^d, C_{s}^d) = ConvLSTM_D(M_s^d, H_{s-1}^d,C_{s-1}^d)
\label{eq:refine}
\end{equation}
where $H_{s}^d$ represents the hidden state of each time step over different level $s$ in the refinement. The hidden state $H_{S}^d$ at the last time step of Eq.~\ref{eq:refine} is adopted as the output of the decoder network. Finally, $H_{S}^d$ is fed to another convolutional layer to produce a 2-D probability score map $y\in(0,1)$ normalized with sigmoid function. The probability score can be trained with a ground truth label map $\hat{y}$ by a binary cross entropy loss function as:
\begin{equation}
    Loss = -\frac{1}{\Omega}\sum_{n=1}^{\Omega}(\hat{y}(n)\log y(n)+(1-\hat{y}(n))\log(1-y(n)))
\end{equation}
where $\Omega$ is the whole set of pixels in the image and $n$ is n-th pixel in it.


\begin{table*}[ht!]
\begin{center}
\caption{Ablation study of the relative contributions of different components of the proposed network on the UNC val set. The first three rows correspond to variants of our model where some components are removed. The 4th row is our model. The last row is our model with DCRF postprocessing. The results show that each component of our model help improving the performance.}

\scalebox{1.15}{
\begin{tabular}{l||c|c|c|c|c|c}
\hline
    Method &  prec@0.5 & prec@0.6 &  prec@0.7  & prec@0.8 &  prec@0.9 & IoU\\
\hline
    E-ConvLSTM(w/o word attention) & 47.15 & 36.97 & 25.63 & 12.84 & 2.14 & 46.70  \\
    E-ConvLSTM   & 54.62  & 44.20 & 30.77 & 16.02 & 2.56 & 50.50   \\
    E-ConvLSTM + D-ConvLSTM(w/o spatial attention)  & 65.54  & 57.27 & 46.82 & 30.54 & 8.65 & 56.27   \\
    E-ConvLSTM + D-ConvLSTM                  & 68.67  & 60.95 & 50.48 & 32.96 & 9.37 & 58.62   \\
    E-ConvLSTM + D-ConvLSTM + DCRF    & 68.97 & 61.49 & 51.98 & 36.15 & 11.42 & 59.04\\
\hline
\end{tabular}
}
\end{center}
\label{tab:ab}
\end{table*}

\begin{table*}[ht!]
\begin{center}
\caption{Comparison of the segmentation performance with the state-of-the-art methods in term of IoU. The top five methods are directly evaluated by taking the outputs of networks and the bottom three results are post-processed with DCRF.
}

\scalebox{1.1}{
\begin{tabular}{c||c|ccc|ccc|c}
\hline
	& Google-Ref &  &UNC& & &UNC+&   & Referit\\

	  & val   & val & testA & testB &  val & testA & testB  & test\\
\hline
\hline
RMI~\cite{liu2017recurrent} & 34.40 & 44.33 & 44.74 & 44.63  & 29.91 & 30.37 & 29.43   & 57.34 \\
DMN~\cite{margffoy2018dynamic}  & 36.76 & 49.78 & 54.83 & 45.13 & 38.88 & 44.22 & 32.29   & 52.81 \\
KWA~\cite{shi2018key}  & 36.92 & - & - & -  & - & - & -   & 59.09 \\
RRN~\cite{li2018referring}   & 36.32    & 54.26 & 56.21 & 52.71 & 39.23 & 41.68 & 35.63  & 63.12 \\
CMSA~\cite{ye2019cross}   & 39.87   & 58.00 & \textbf{60.33} & 54.74 & 43.57 & 47.27 &  37.73 & 63.17 \\
Ours		& \textbf{41.32} &  \textbf{58.62} & 59.73 &  \textbf{56.23}      &   \textbf{44.18} & \textbf{47.44} & \textbf{39.43}           & \textbf{63.75}  \\
\hline
RMI + DCRF~\cite{liu2017recurrent} & 34.52 & 45.18 & 45.69 & 45.57 & 29.86 & 30.48 & 29.50   & 58.73 \\
RRN + DCRF~\cite{li2018referring}   & 36.45    & 55.33 & 57.26 & 53.93 & 39.75 & 42.15 & 36.11  & 63.63 \\
CMSA +DCRF~\cite{ye2019cross}	& 39.98   & 58.32 & 60.61 & 55.09      &   43.76 & 47.60 & 37.89          & 63.80  \\
Ours + DCRF	& \textbf{41.77}	& \textbf{59.04} & \textbf{60.74} & \textbf{56.73}      &   \textbf{44.54} & \textbf{47.92} & \textbf{39.73}          & \textbf{63.92}  \\
\hline
\end{tabular}
}
\end{center}
\label{tab:miou}
\end{table*}

\section{Experimental Setup}\label{sec:experiment}
In this section, we introduce the datasets in Sec.~\ref{sec:datasets} and evaluation metrics in Sec.~\ref{sec:eval}, and also describe the implementation details of the proposed approach in Sec.~\ref{sec:implement}.

\subsection{Datasets}\label{sec:datasets}
We evaluate our model on four publicly available datasets including Google-Ref~\cite{mao2016generation}, UNC~\cite{yu2016modeling}, UNC+~\cite{yu2016modeling} and Referit~\cite{kazemzadeh2014referitgame}.  All experiments are conducted under the same train and test split sets as~\cite{liu2017recurrent}.

The Google-Ref dataset is composed of 104,560 expressions referring, 54,822 objects and 26,711 images. These images and ground truth masks are collected from the MS COCO dataset~\cite{lin2014microsoft} and referring expressions are annotated from Amazon Mechanical Turk. The referring expressions of the Google-Ref dataset are longer with an average length of 8.43 words compared with the other three datasets and multiple objects with the same category can appear in a single image.

The UNC dataset is also based on the MS COCO dataset which contains 19,994 images with 142,209 referring expressions for 50,000 objects. It is gathered by a two-player game~\cite{kazemzadeh2014referitgame} where one player annotates the image region according to the expression by other player interactively. 
Both location and appearance words can be used to describe the referred objects.

The UNC+ dataset is similar as the UNC dataset  with a total of 141,564 expressions for 49,856 objects in19,992 images. The main difference from the UNC dataset is that location information is not allowed to refer to the object of interest. In other words, referring expressions purely rely on the appearance and context descriptions to describe referred objects.

The Referit dataset is built on IAPR TC-12 dataset~\cite{escalante2010segmented} which includes masks for stuffs, e.g., ``sky'' and ``ground'', in addition to objects. Same as the UNC and UNC+ datasets,  referring expressions of   Referit is also collected by the two-player game. It has 130,525 expressions referring to 96,654 object masks in 19,894 images in total.

\subsection{Evaluation Metrics}\label{sec:eval}
Following the previous work~\cite{liu2017recurrent}, we use Intersection-over-union (IoU) and Precision@X (Prec@X) as the evaluation metrics. IoU measures the intersection area divides by the union area between ground-truth and a predicted segmentation mask averaged over all test data. To be more specific, let $S$ and $S'$ be the ground-truth and predicted segmentation masks, respectively. IoU is defined as the ratio between the intersection and the union of these two masks, i.e. $\frac{S\cap S'}{|S\cup S'|}$. 
For a more precise comparison, Prec@X is provided to evaluate detailed contributions in ablation study. It measures the percentage of test images which have higher IoU than a threshold $X$. We choose the value of $X$ ranging from $0.5$ to $0.9$ with an interval of $0.1$.

\subsection{Implementation Details}\label{sec:implement}
We use DeepLab-101 network~\cite{chen2016deeplab} with pre-trained weights from the Pascal VOC dataset~\cite{everingham2012pascal} as previous works~\cite{liu2017recurrent,shi2018key,li2018referring,ye2019cross} to extract visual features. The input image is resized and zero-padded to $320\times320$ and keep the maximum length of the referring expression as $20$, so the spatial dimensions of the visual features from \{$Res5, Res4, Res3$\} are the same as $W=H=40$ thanks to dilated convolution. The visual features at different levels are transformed to a fixed channel size of $C_v =1000$ using $1\times 1$ convolution. 
Every word is first embedded to a $1000$ dimensional vector and  then passed through a bidirectional LSTM with the cell size of $500$. After combining both forward and backward hidden outputs, the dimension of the language feature for every word is $C_r=1000$.
The cell sizes of E-ConvLSTM and D-ConvLSTM are set to be $1000$ and $500$, respectively. Furthermore, we apply DCRF~\cite{krahenbuhl2011efficient} which is a widely used post-processing operation for precise segmentation masks.
In order to minimize the loss function, the network is trained with Adam optimization algorithm~\cite{kingma2014adam} with initial learning rate of $0.00025$, weight decay of $0.0005$. We employ a ``poly'' strategy~\cite{chen2016deeplab} to adaptively tune the learning rate with power of $0.9$.

\begin{figure}[htb]
  \centering
  \includegraphics[width=8cm]{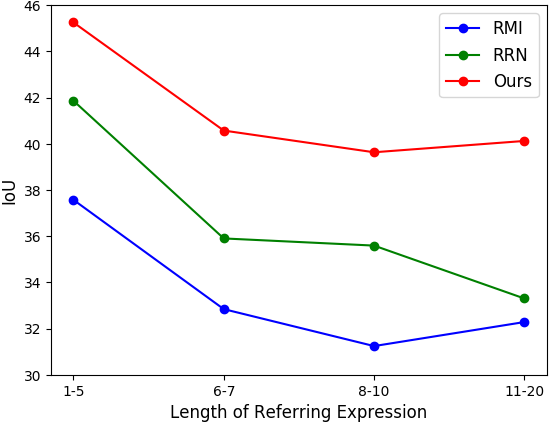}
\caption{Comparison of segmentation performance of different lengths of referring expressions on Google-Ref. }
\label{fig:len}
\end{figure}

\begin{figure*}[htb]
  \centering
{\setlength{\tabcolsep}{0.5pt}
\scalebox{0.7}{
  \begin{tabular}{ccccccc}

    \multicolumn{7}{c}{Referring expression: \textit{``the white car''}} \\
    \includegraphics[width=1.2in]{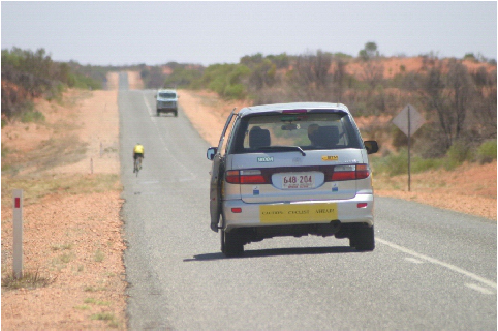}&
    \includegraphics[width=1.2in]{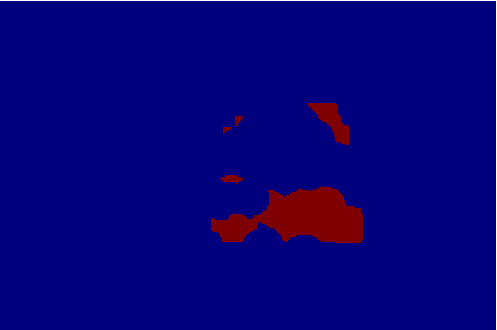}&
    \includegraphics[width=1.2in]{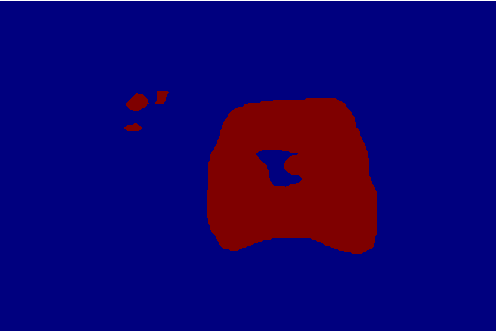}&
    \includegraphics[width=1.2in]{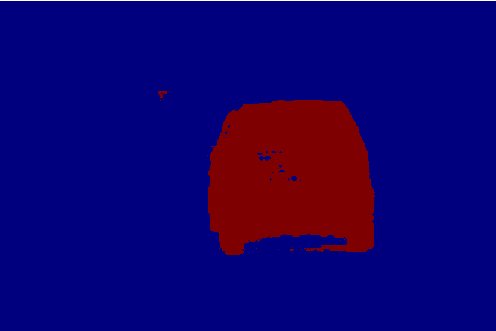}&
      \includegraphics[width=1.2in]{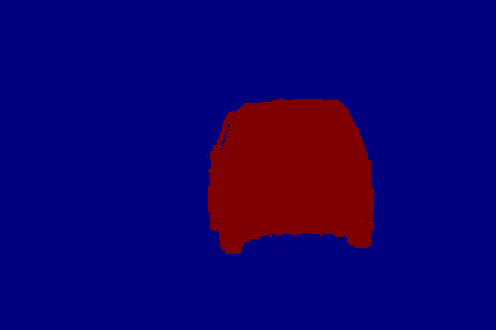}&
     \includegraphics[width=1.2in]{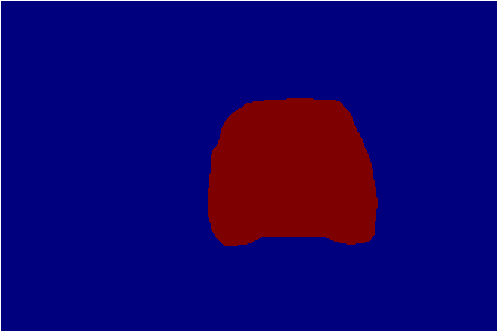}&
      \includegraphics[width=1.2in]{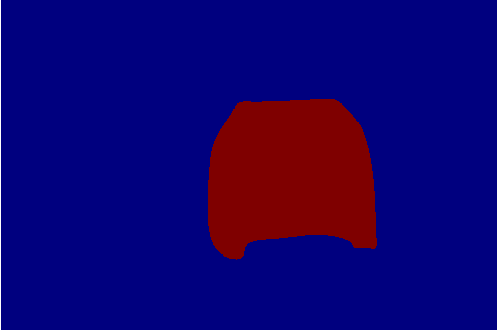}\\
        
    \multicolumn{7}{c}{Referring expression: \textit{``brown cow''}} \\
    \includegraphics[width=1.2in]{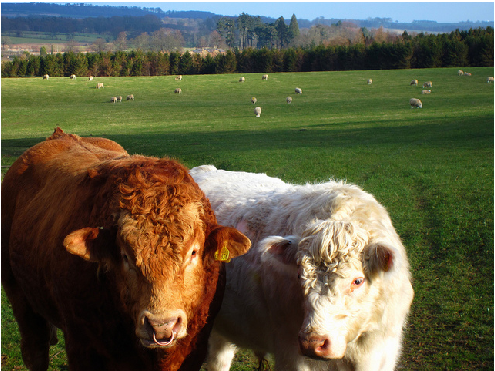}&
    \includegraphics[width=1.2in]{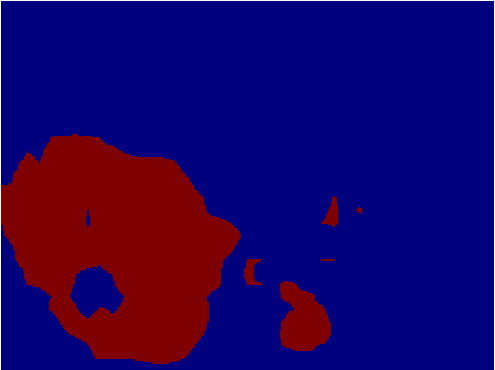}&
    \includegraphics[width=1.2in]{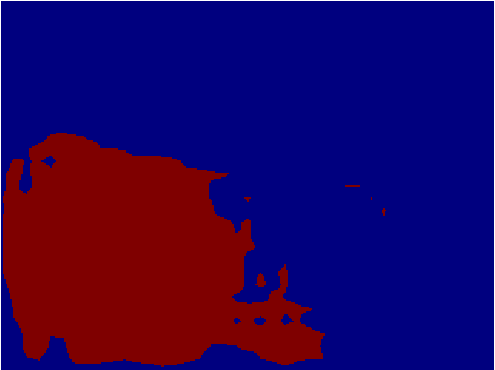}&
      \includegraphics[width=1.2in]{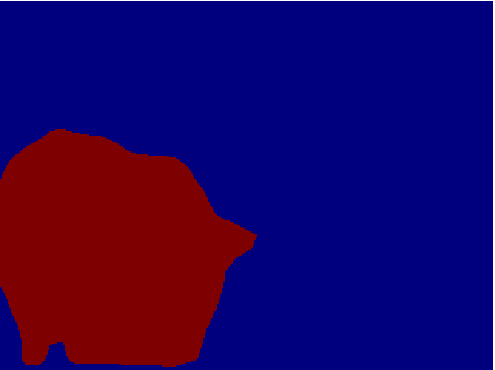}&
       \includegraphics[width=1.2in]{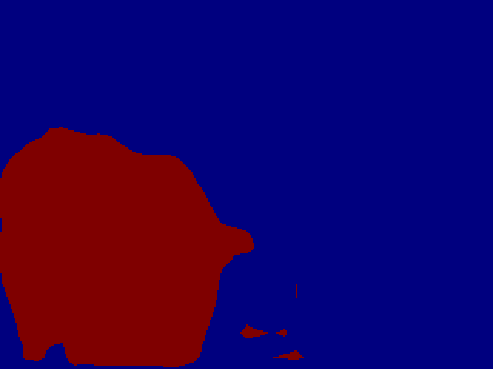}&
     \includegraphics[width=1.2in]{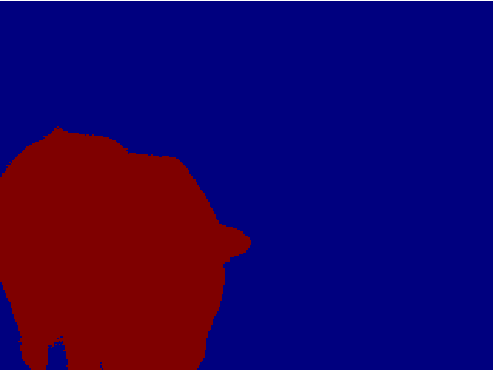}&
      \includegraphics[width=1.2in]{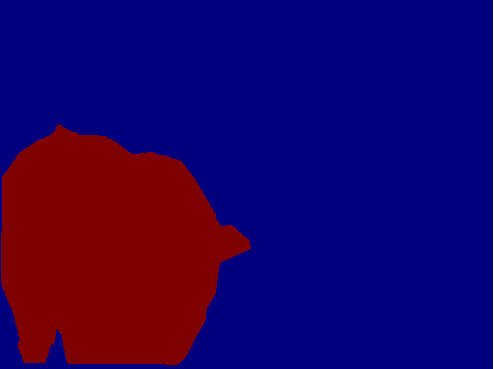}\\

        \multicolumn{7}{c}{Referring expression: \textit{``smaller container''}} \\
    \includegraphics[width=1.2in]{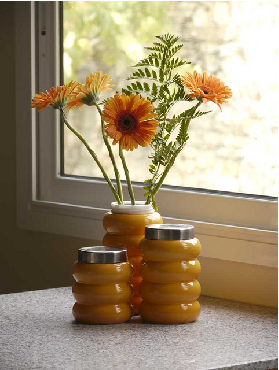}&
    \includegraphics[width=1.2in]{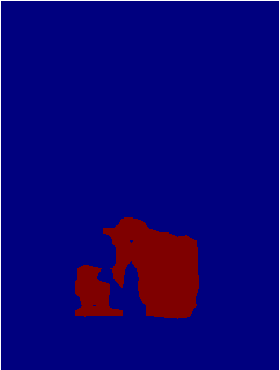}&
    \includegraphics[width=1.2in]{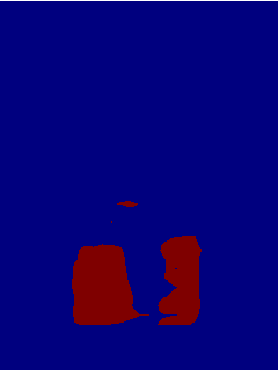}&
    \includegraphics[width=1.2in]{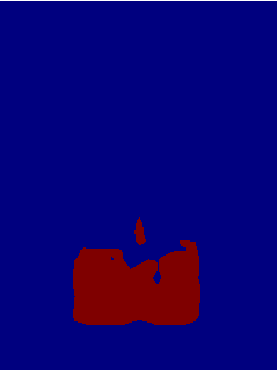}&
     \includegraphics[width=1.2in]{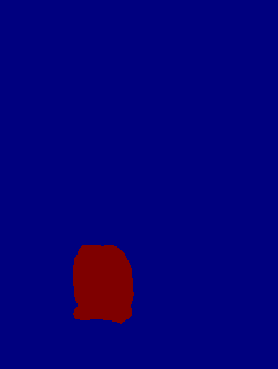}&
     \includegraphics[width=1.2in]{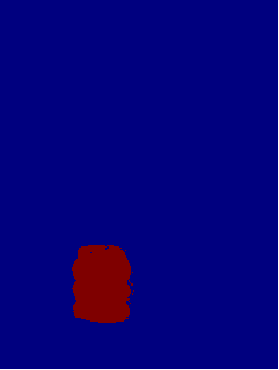}&
      \includegraphics[width=1.2in]{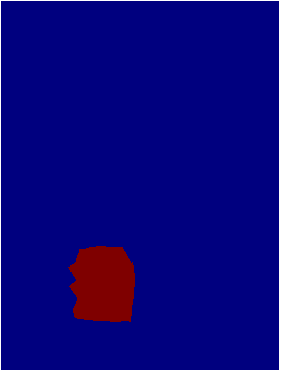}\\
    
        \multicolumn{7}{c}{Referring expression: \textit{``portion of building right of girl's head''}} \\
    \includegraphics[width=1.2in]{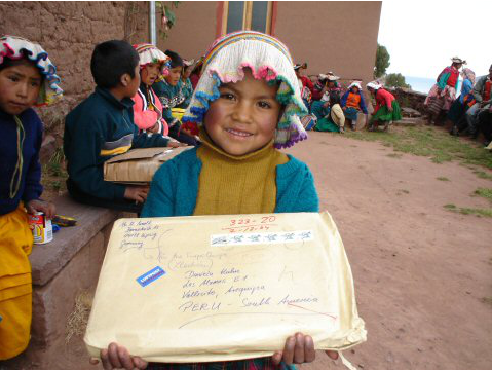}&
    \includegraphics[width=1.2in]{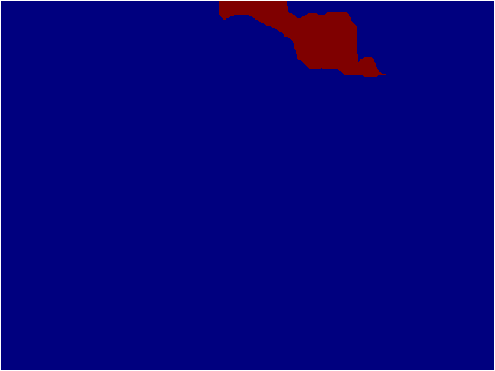}&
    \includegraphics[width=1.2in]{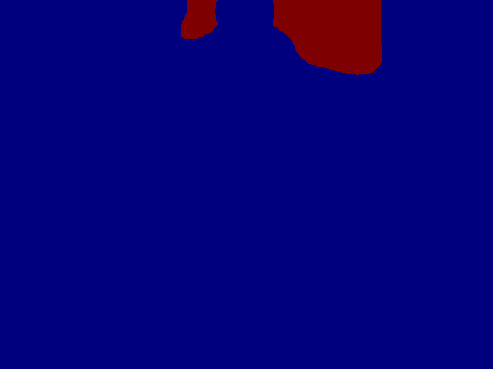}&
    \includegraphics[width=1.2in]{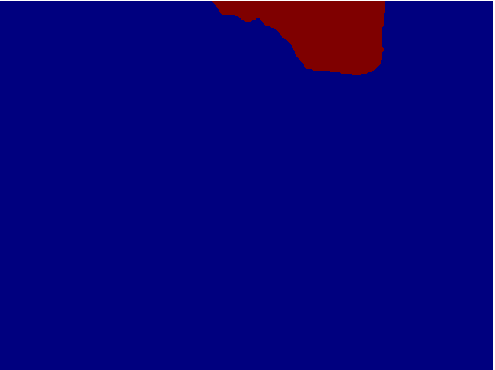}&
     \includegraphics[width=1.2in]{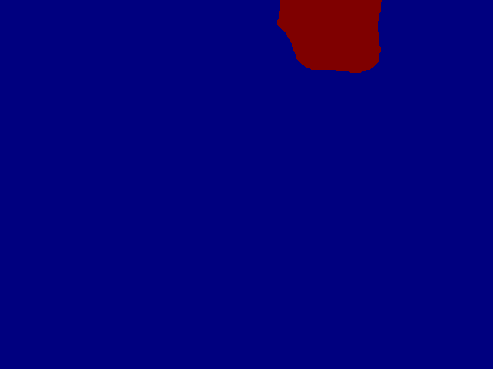}&
     \includegraphics[width=1.2in]{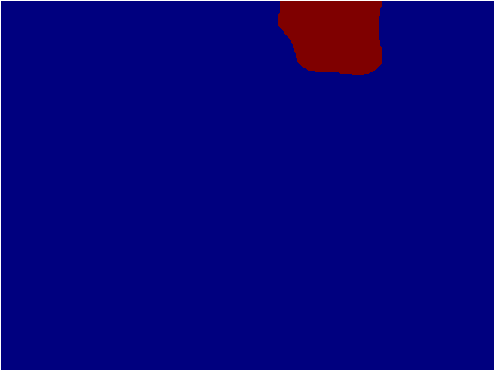}&
      \includegraphics[width=1.2in]{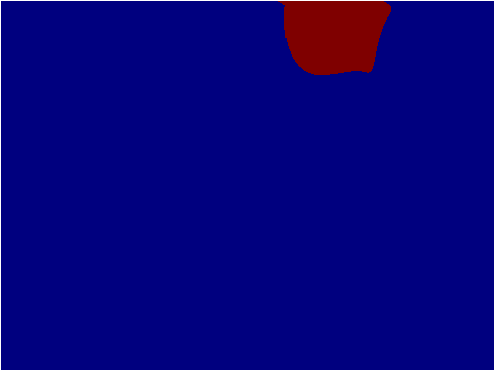}\\
  
       \multicolumn{7}{c}{Referring expression: \textit{``a black bowl of vegetable stir fry''}} \\
    \includegraphics[width=1.2in]{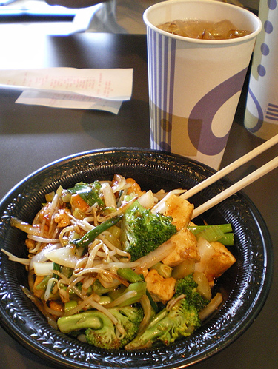}&
    \includegraphics[width=1.2in]{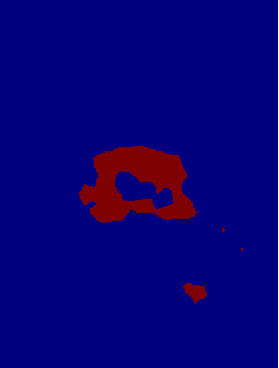}&
    \includegraphics[width=1.2in]{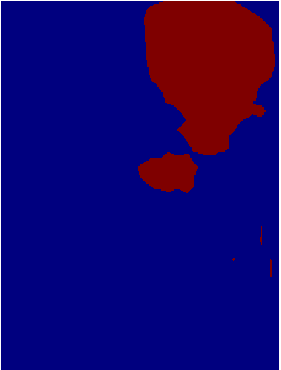}&
    \includegraphics[width=1.2in]{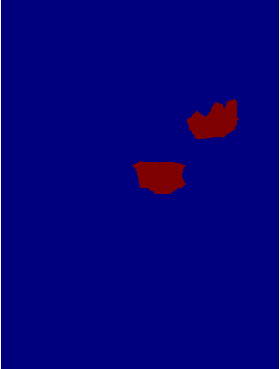}&
     \includegraphics[width=1.2in]{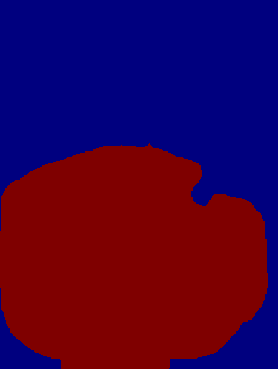}&
     \includegraphics[width=1.2in]{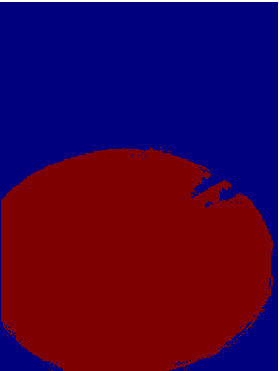}&
      \includegraphics[width=1.2in]{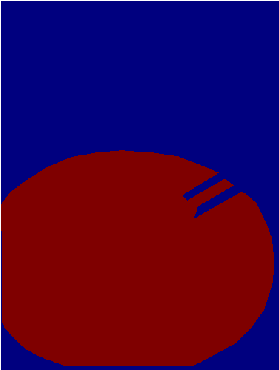}\\   
        
        \multicolumn{7}{c}{Referring expression: \textit{``sky''}} \\
    \includegraphics[width=1.2in]{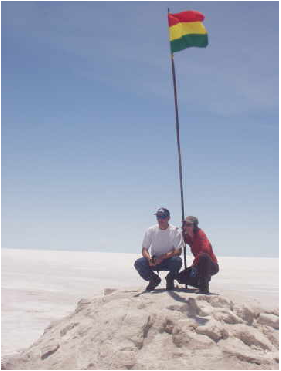}&
    \includegraphics[width=1.2in]{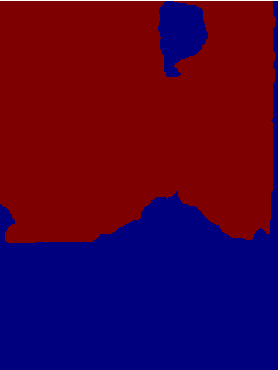}&
    \includegraphics[width=1.2in]{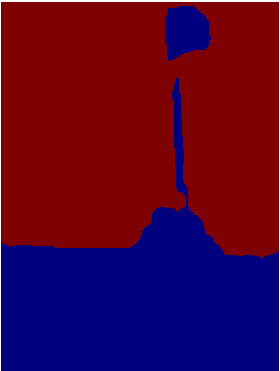}&
    \includegraphics[width=1.2in]{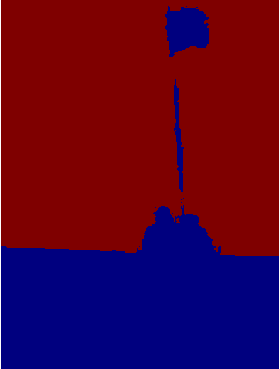}&
     \includegraphics[width=1.2in]{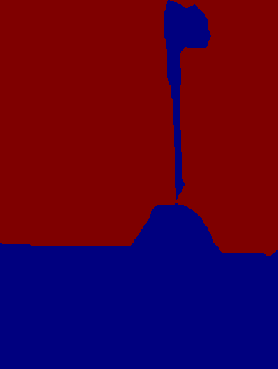}&
     \includegraphics[width=1.2in]{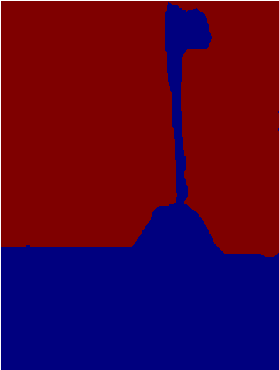}&
      \includegraphics[width=1.2in]{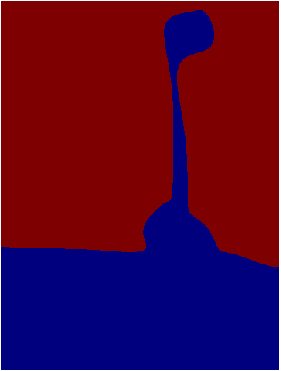}\\
      
    (a) & (b) & (c) & (d) & (e) & (f) & (g)\\
  \end{tabular}
}
}

\caption{
Visual comparison of referring image segmentation results: (a) original image; (b,c,d,e,f) referring image segmentation masks from \cite{liu2017recurrent}, \cite{li2018referring}, \cite{ye2019cross} and our network without or with DCRF; (g) ground-truth segmentation mask. The corresponding referring expressions are shown above these images in each row.}
\label{fig:qual}
\end{figure*}

\begin{figure*}[htb]
  \centering
{\setlength{\tabcolsep}{0.5pt}
\scalebox{0.74}{
  \begin{tabular}{cccccc}

    \multicolumn{6}{c}{\includegraphics[height=0.45in]{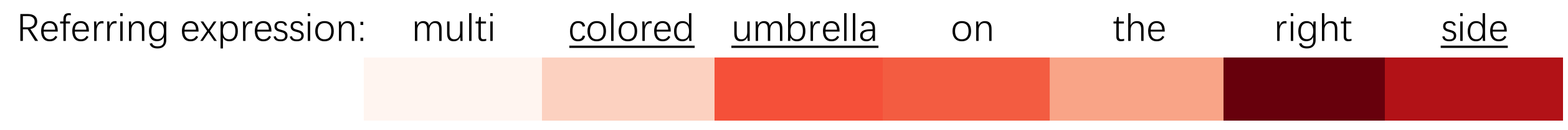}}\\
    \includegraphics[width=1.2in]{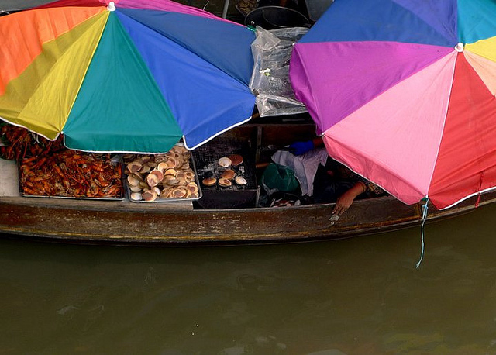}&
    \includegraphics[width=1.2in]{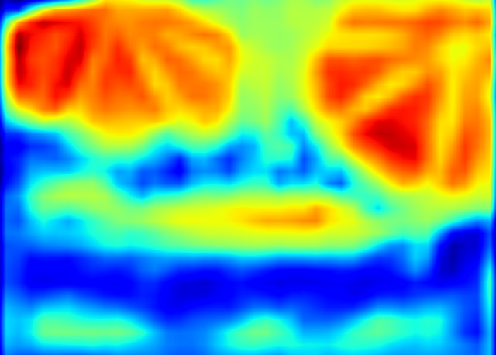}&
    \includegraphics[width=1.2in]{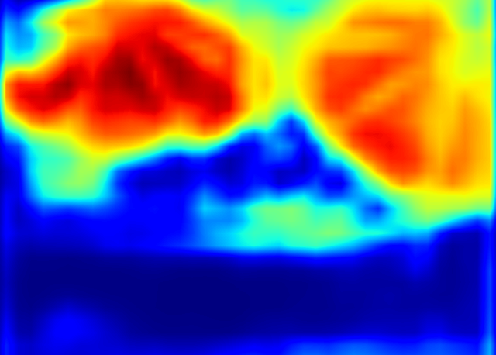}&
    \includegraphics[width=1.2in]{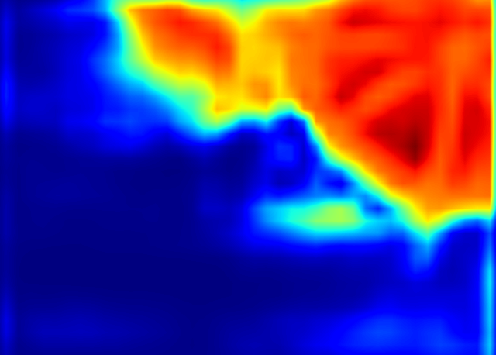}&
     \includegraphics[width=1.2in]{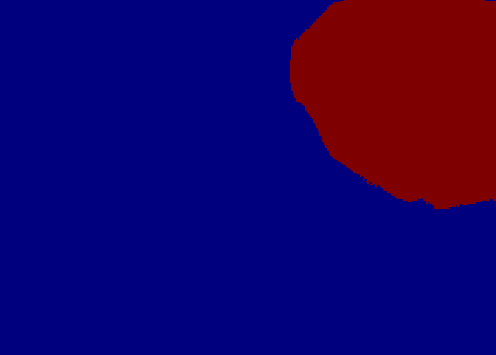}&
      \includegraphics[width=1.2in]{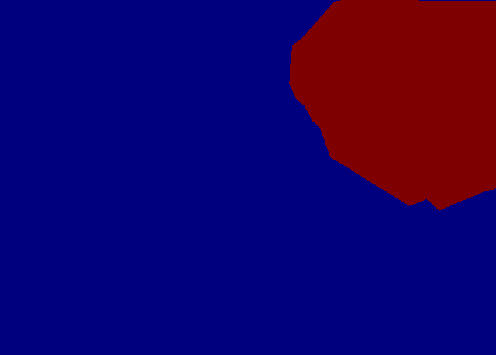}\\
        
    \multicolumn{6}{c}{\includegraphics[height=0.45in]{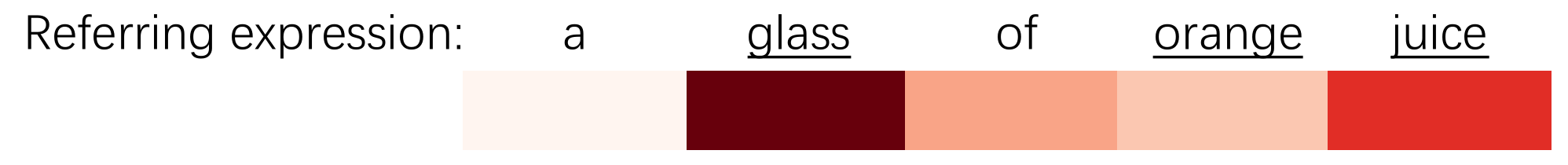}}\\
    \includegraphics[width=1.2in]{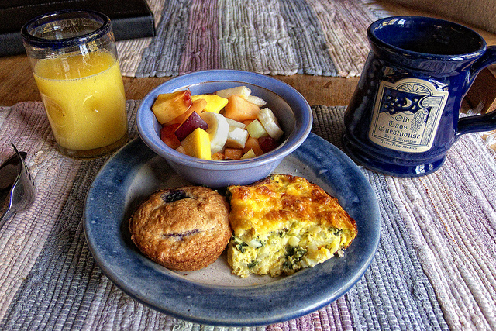}&
    \includegraphics[width=1.2in]{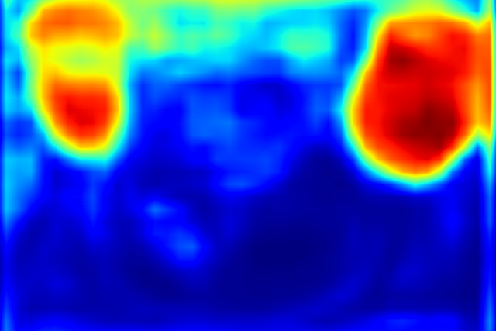}&
    \includegraphics[width=1.2in]{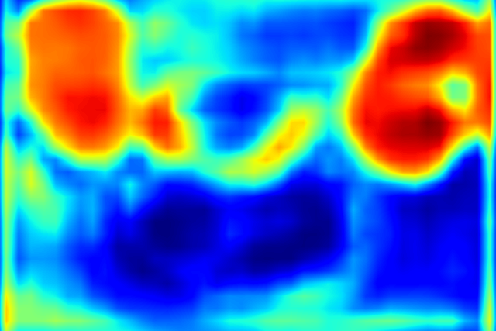}&
     \includegraphics[width=1.2in]{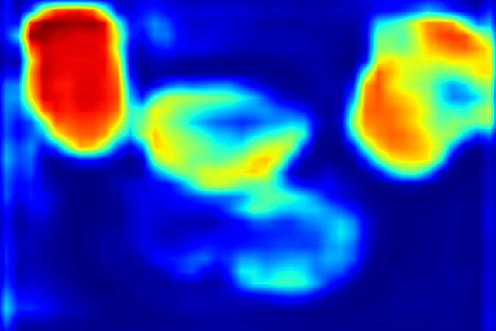}&
     \includegraphics[width=1.2in]{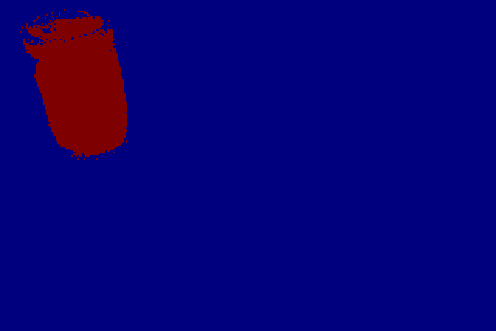}&
      \includegraphics[width=1.2in]{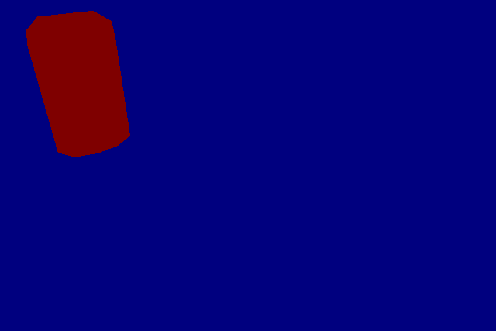}\\
      (a) & (b) & (c) & (d) & (e) & (f)\\
    
  \end{tabular}
}
}

\caption{Visualization on how E-ConvLSTM works for multimodal interaction: (a) original image; (b,c,d) intermediate activation by mean-pooling hidden state output after each underline word has been processed by the encoder network; (e) predicted segmentation mask; (f) ground truth segmentation mask.}
\label{fig:word}
\end{figure*}

\begin{figure*}[htb]
  \centering
{\setlength{\tabcolsep}{0.5pt}
\scalebox{0.74}{
  \begin{tabular}{cccccc}

    \multicolumn{6}{c}{\includegraphics[height=0.45in]{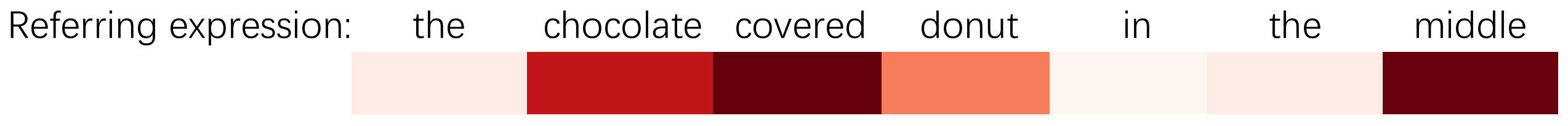}}\\
    \includegraphics[width=1.2in]{figure/fig_3l/1452_im.png}&
    \includegraphics[width=1.2in]{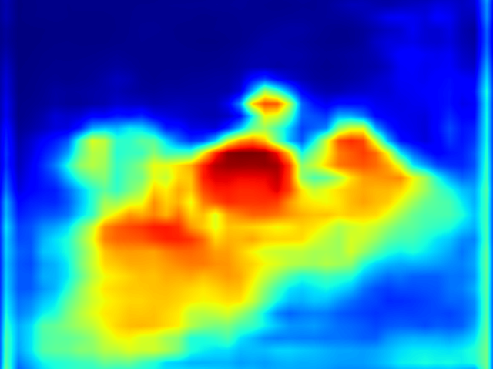}&
    \includegraphics[width=1.2in]{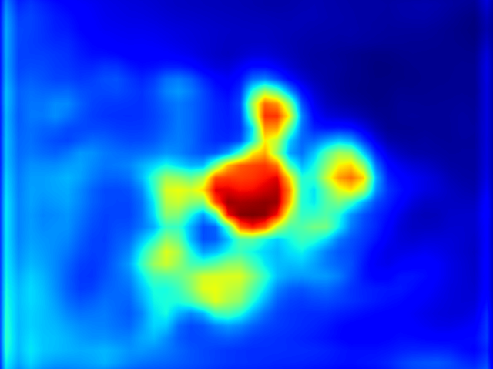}&
     \includegraphics[width=1.2in]{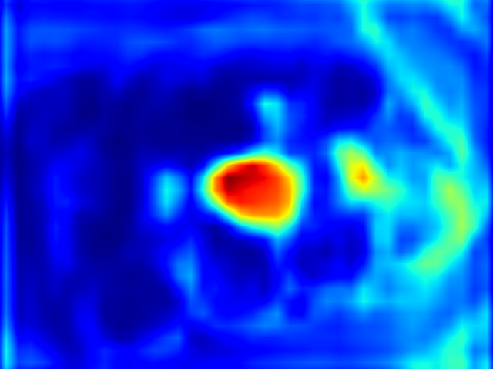}&
     \includegraphics[width=1.2in]{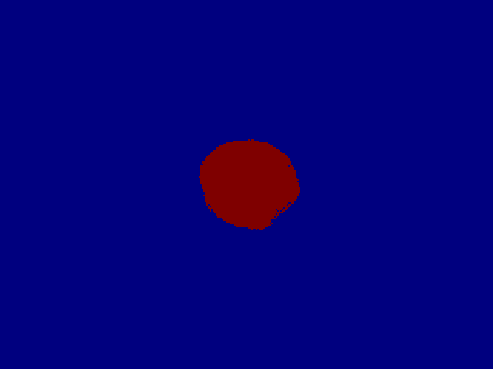}&
      \includegraphics[width=1.2in]{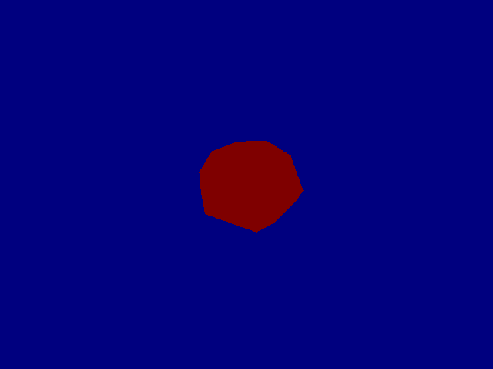}\\
        
    \multicolumn{6}{c}{\includegraphics[height=0.45in]{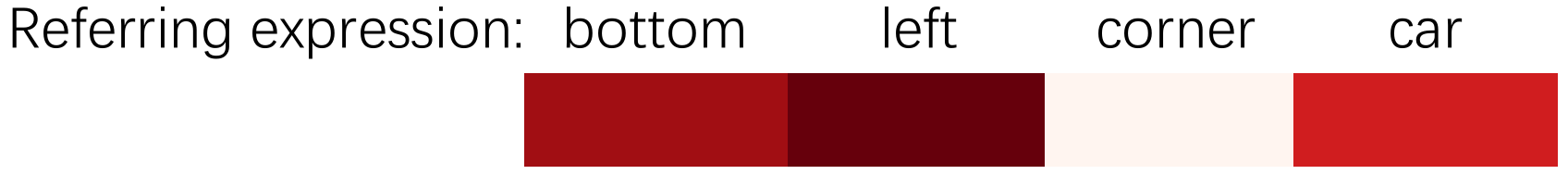}}\\
    \includegraphics[width=1.2in]{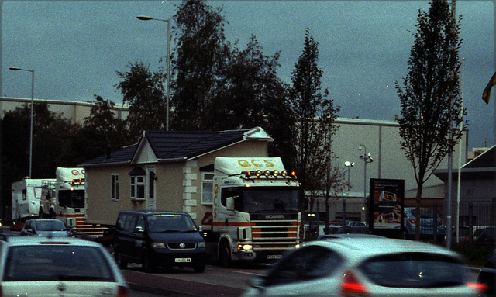}&
    \includegraphics[width=1.2in]{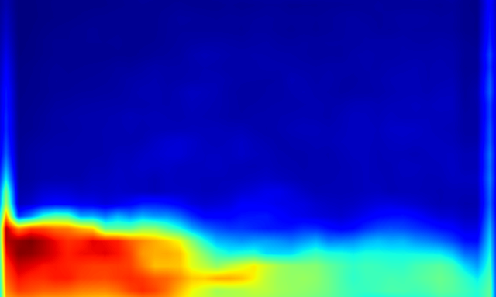}&
    \includegraphics[width=1.2in]{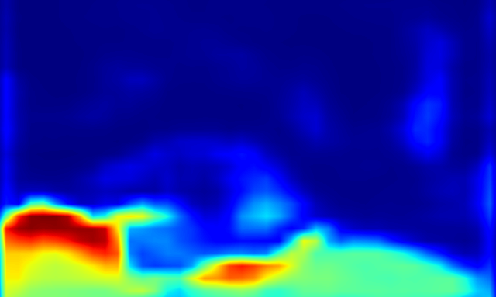}&
     \includegraphics[width=1.2in]{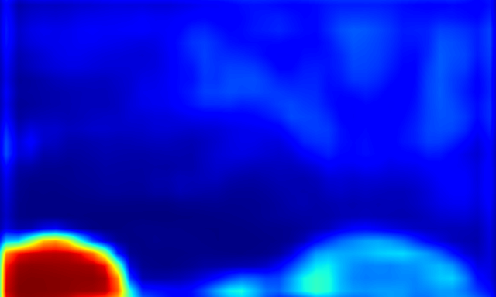}&
     \includegraphics[width=1.2in]{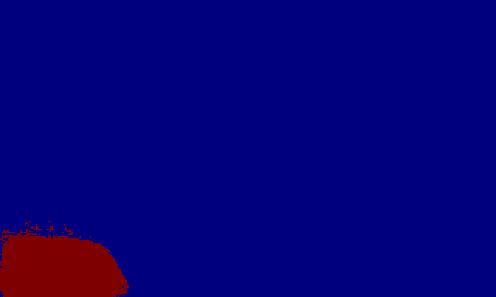}&
      \includegraphics[width=1.2in]{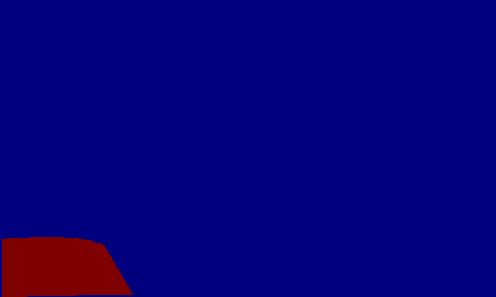}\\
    (a) & (b) & (c) & (d) & (e) & (f)\\
  \end{tabular}
}
}

\caption{
Visualization to illustrate the hidden states of D-ConvLSTM after refining with low-level features gradually: (a) original image; (b,c,d) intermediate activation by mean-pooling hidden state outputs; (e) predicted segmentation mask; (f) ground-truth segmentation mask.}
\label{fig:mul}
\end{figure*}

\begin{figure*}[!htb]
  \centering
{\setlength{\tabcolsep}{0.5pt}
\scalebox{0.9}{
  \begin{tabular}{cccc}

     & & \textit{``a chair to the} &  \\
     & \textit{``chair"} & \textit{right of a woman"}      &\textit{``a right chair is empty"} \\%
        \includegraphics[width=1.21in]{figure/fig_words_vis/3934_im.png}&
    \includegraphics[width=1.21in]{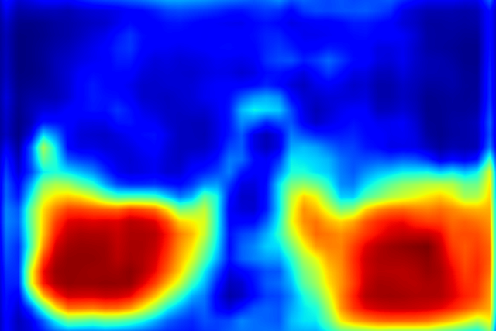}&
    \includegraphics[width=1.21in]{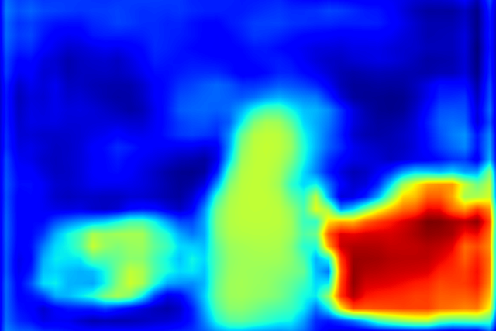}& 
     \includegraphics[width=1.21in]{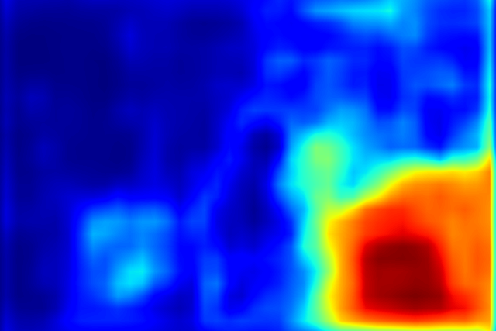}\\
     
       & & & \textit{``a girl in a yellow shirt} \\
       &  \textit{``girl"} & \textit{``girl on the grass"} &\textit{sitting on the grass"} \\
       \includegraphics[width=1.21in]{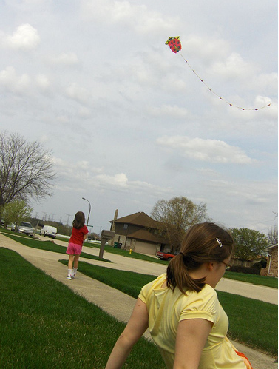}&
    \includegraphics[width=1.21in]{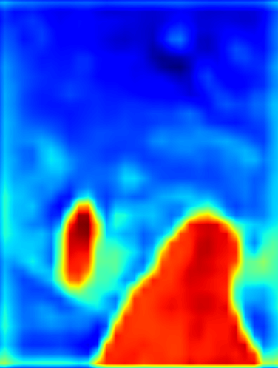}&
    \includegraphics[width=1.21in]{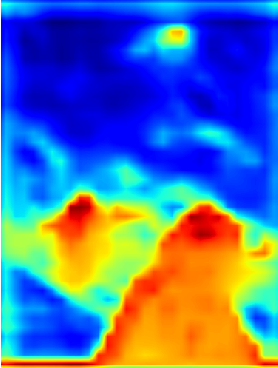}& 
     \includegraphics[width=1.21in]{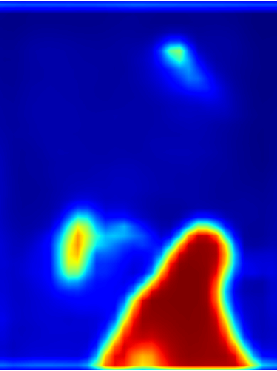}\\

 \end{tabular}
}
}
\caption{Visualization of the feature representation. These spatial heatmaps show the responses of our network to the diverse referring expressions that are not included in the dataset.}
\label{fig:words_vis}
\end{figure*}

\begin{figure}[htb]
  \centering
{\setlength{\tabcolsep}{0.5pt}
\scalebox{0.85}{
  \begin{tabular}{cccccc}

    \multicolumn{4}{c}{Referring expression: \textit{``broccoli plate back left''}} \\
    \includegraphics[width=1.0in]{figure/fig_comp_cap/3296_im}&
    \includegraphics[width=1.0in]{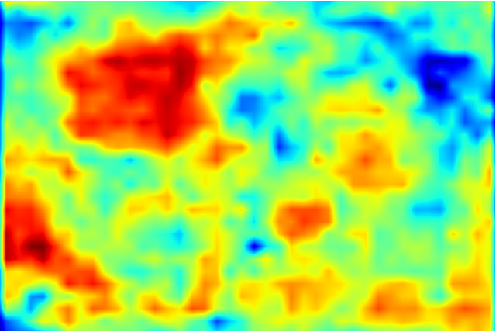}&
    \includegraphics[width=1.0in]{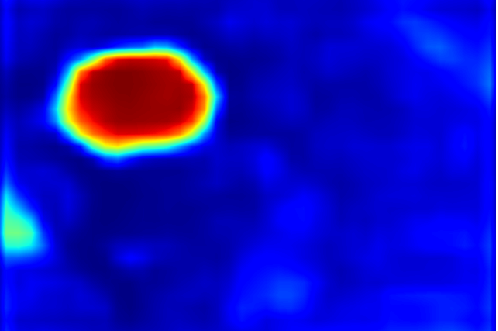}&
      \includegraphics[width=1.0in]{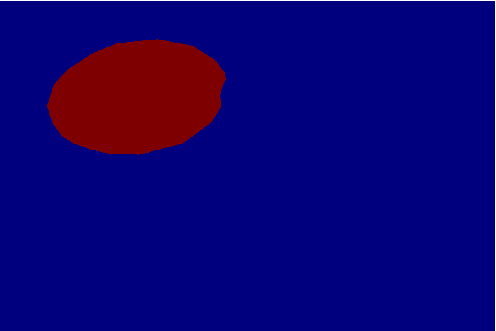}\\
      
       \multicolumn{4}{c}{Referring expression: \textit{``far left front guy''}} \\
    \includegraphics[width=1.0in]{figure/fig_comp_cap/4855_im}&
    \includegraphics[width=1.0in]{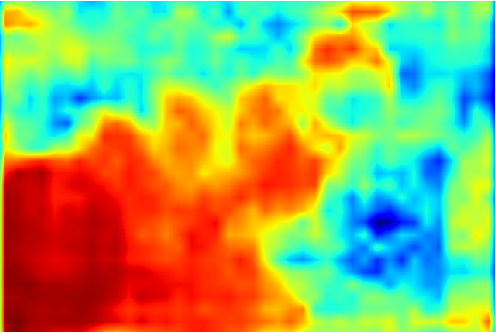}&
    \includegraphics[width=1.0in]{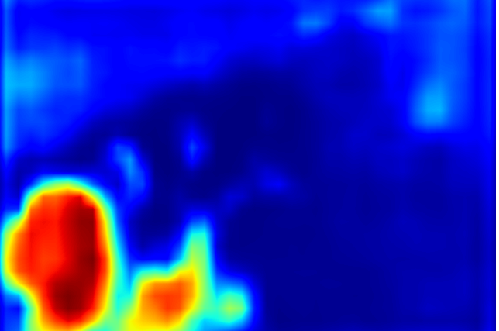}&
      \includegraphics[width=1.0in]{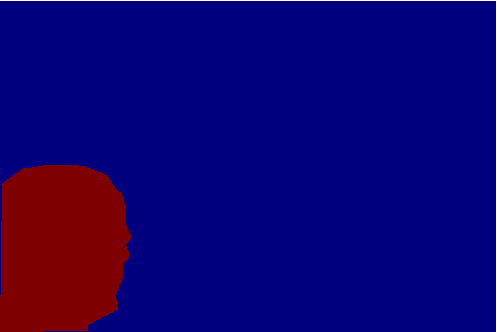}\\

     \multicolumn{4}{c}{Referring expression: \textit{``guys shoulder on right''}} \\
     \includegraphics[width=1.0in]{figure/fig_comp_cap/3578_im}&
    \includegraphics[width=1.0in]{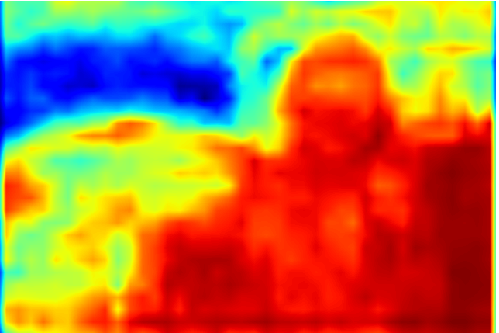}&
    \includegraphics[width=1.0in]{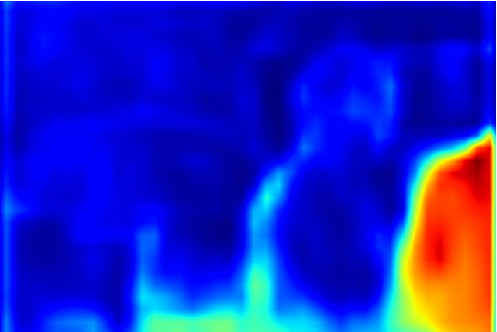}&
      \includegraphics[width=1.0in]{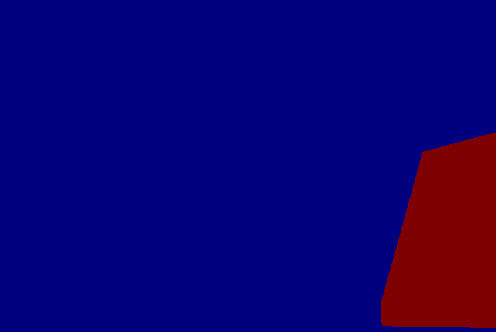}\\
      
      (a) & (b) & (c) & (d)\\
    
  \end{tabular}
}
}

\caption{Comparison of visualization results of our attention model with the attention method used in image captioning. (a) original image; (b) visualized result of the attention model in~\cite{xu2015show}; (c) visualized result of our E-ConvLSTM; (d) ground truth segmentation mask.}
\label{fig:comp_cap}
\end{figure}

\begin{figure}[htb]
  \centering
{\setlength{\tabcolsep}{0.5pt}
\scalebox{0.85}{
  \begin{tabular}{cccccc}

    \multicolumn{4}{c}{Referring expression: \textit{``bottom hotdog''}} \\
    \includegraphics[width=1.0in]{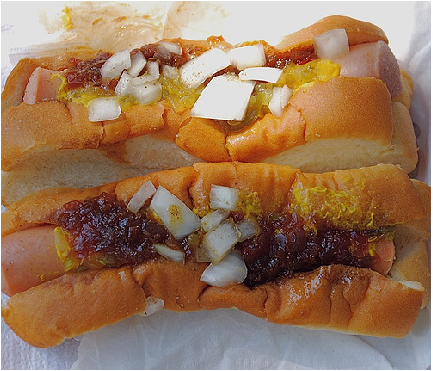}&
    \includegraphics[width=1.0in]{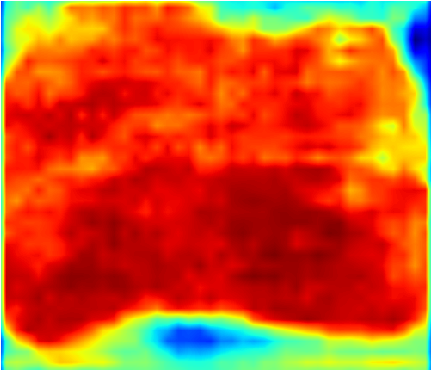}&
    \includegraphics[width=1.0in]{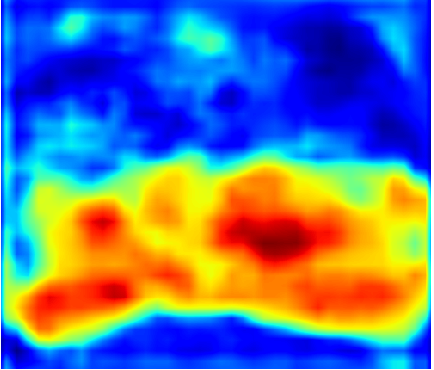}&
      \includegraphics[width=1.0in]{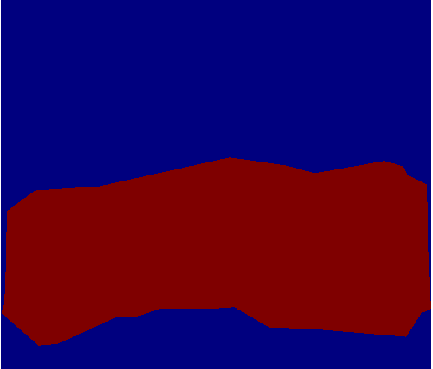}\\
      
       \multicolumn{4}{c}{Referring expression: \textit{``chocolate donut second from right in front row''}} \\
    \includegraphics[width=1.0in]{figure/fig_wo_sp/3049_im.png}&
    \includegraphics[width=1.0in]{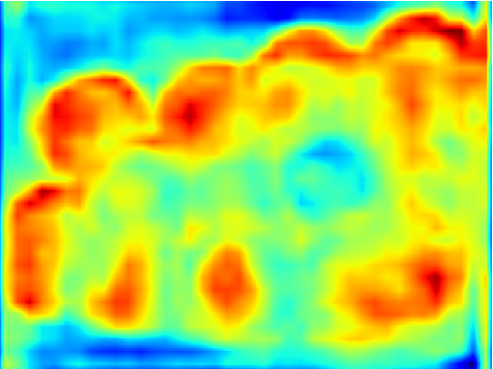}&
    \includegraphics[width=1.0in]{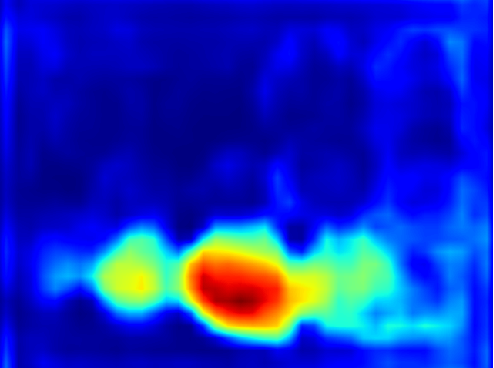}&
      \includegraphics[width=1.0in]{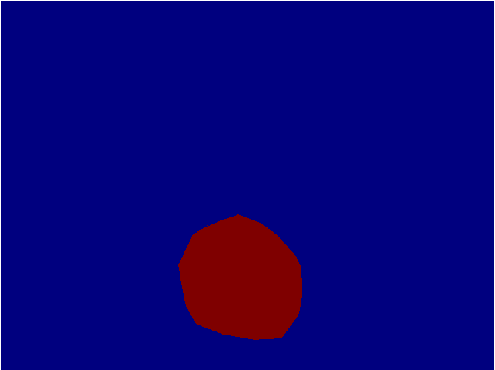}\\

      (a) & (b) & (c) & (d)\\
    
  \end{tabular}
}
}

\caption{Comparison of visualization results of our network with or without the spatial feature. (a) original image; (b) visualized result without spatial features; (c) visualized result with spatial features; (d) ground truth segmentation mask.}
\label{fig:wo_sp}
\end{figure}

\section{Performance Evaluation}\label{sec:result}
We perform ablation study in Sec.~\ref{sec:ablation} to evaluate the relative contributions of various components of the proposed network. We also present quantitative and qualitative results for comparisons with other state of the art methods in Sec.~\ref{sec:quan}  and  Sec.~\ref{sec:qual}, respectively.

\subsection{Ablation Study}\label{sec:ablation}
To verify the effectiveness of each component of our dual convolutional LSTM network, we first conduct ablation experiments on the UNC validation dataset with the following different variants of the proposed method.
\begin{itemize}
\item E-ConvLSTM(w/o word attention): This model does not use multi-level segment decoder to iteratively refinement of the segmentation mask. Instead the segmentation mask is directly predicted from the output of the multimodal feature encoder. This model also does not use word attentions.
\item E-ConvLSTM: This is similar to the previous model, but uses word attentions to adaptively encode the multimodal interaction.
\item E-ConvLSTM + D-ConvLSTM(w/o spatial attention): This is similar to the proposed model, but without the spatial attentions for the multimodal feature maps.
\item E-ConvLSTM + D-ConvLSTM: This is the complete proposed model.
\item E-ConvLSTM + D-ConvLSTM + DCRF: This is our proposed model with DCRF post-processing.
\end{itemize}

The results of the ablation study are shown in Table~\ref{tab:ab}. It can be seen that introducing word attentions into the cell state of E-ConvLSTM leads to better multimodal interaction result for multimodal feature encoder. In addition, multimodal feature refinement at multiple levels by the multi-level segment decoder considerably improves segmentation performance compared with \textit{encoder-only} methods shown in the first two rows of Table~\ref{tab:ab}. Spatial attention further improves the performance due to a stronger feature representation that focuses on important spatial regions. DCRF achieves more precise segmentation masks in the final result. In summary, each component of our model contributes to improve the segmentation performance.

\subsection{Quantitative Results}\label{sec:quan}
We quantitatively compare our model (with or without DCRF post-processing) with state-of-the-art referring image segmentation methods including RMI~\cite{liu2017recurrent}, DMN~\cite{margffoy2018dynamic}, KWA~\cite{shi2018key}, RRN~\cite{li2018referring} and CMSA~\cite{ye2019cross} in Table \ref{tab:miou}.

The proposed method outperforms all other methods consistently on all datasets with or without DCRF post-processing. Specifically, the performance improvement on Google-Ref is particularly significant. As we mentioned in Sec.~\ref{sec:datasets}, the Google-Ref dataset is much more challenging since it has longer referring expressions and richer descriptions compared with other datasets. 
This demonstrates the importance of the proposed E-ConvLSTM which adaptively focuses on more important words during the multimodal interaction. It plays an important role in understanding the referring expression and localizing the referred object simultaneously. The other methods may fail to capture the long-range dependency since considering each word equally such as RMI~\cite{liu2017recurrent} and RRN~\cite{li2018referring} or without sequentially multimodal interaction such as KWA~\cite{shi2018key} and CMSA~\cite{ye2019cross}. 
In addition, the approaches with multi-level refinement e.g., RRN~\cite{li2018referring}, CMSA~\cite{ye2019cross} and our model achieve better results than other compared methods. It demonstrates that multi-level features help segment objects with clear boundaries and our model benefits from the D-ConvLSTM decoder with spatial attentions.

In order to clearly show the effect of the length of referring expressions for different models, we follow \cite{liu2017recurrent} and evaluate the performance separately for four different groups of the referring expression length range at [1-5], [6-7], [8-10] and [11-20]. As illustrated in Fig.~\ref{fig:len}, our model outperforms other methods in all cases. The performance gain is particularly noticeable with longer referring expressions. This also manifests the advantage of the proposed E-ConvLSTM encoder which embeds word attentions in the ConvLSTM cell to adaptively encode the multimodal interaction in understanding longer and more complicated referring expressions.

\subsection{Qualitative Results}\label{sec:qual}
We show some qualitative examples of the segmentation masks generated by our approach and compare with existing state-of-the-art methods in Fig.~\ref{fig:qual}. 
We can see that our method can successfully handle appearance attributes (the 1st and 2nd examples) where the complete object has internal high contrasts (white car of the 1st example) and low contrasts to the neighboring object (the brown cow and the shady part of nearby cow of 2nd example). The compared methods shown in the column (b), (c), (d) fail to capture the complete objects or suppress the background clearly.
In addition, in the 3rd and 4th examples, our method can accurately identify the objects of interest from complicated referring expressions with relative relationships between homogeneous objects (the container of the 3rd example and the portion of building of the 4th example).  A complete object with heterogeneous components is well-identified in the 5th example and large sky regions are obtained with clear boundaries in the bottom example, while the other methods may not be able to accurately understand the complex referring expressions and obtain precise referred objects. The DCRF post-processing can further refine the segmentation mask as a whole without noisy background regions (the 2nd and 6th examples) and recover precise boundary details (the 3rd and 5th examples).

We also present some visualization examples in Fig.~\ref{fig:word} and Fig.~\ref{fig:mul} to help understand how E-ConvLSTM localizes the referred objects along multimodal interaction and how D-ConvLSTM refines segmentation results with multi-level features, respectively. The reddish blocks under each word presents their word attentions in the expression. The intermediate visualized results are extracted from the hidden state tensors of E-ConvLSTM and D-ConvLSTM, which are meanpooled along channel dimension with normalization and resized to the original image resolution. We show the visualization examples after seen underlined word for Fig.~\ref{fig:word} and after refining features with each level for Fig.~\ref{fig:mul}. It can be observed that E-ConvLSTM can capture the corresponding concepts as far as words have been seen such as ``colored'', ``umbrella'' in the first row and ``glass'' in the second row of Fig.~\ref{fig:word}. Once it interacts with more concrete descriptions in the expression, the network pinpoints to the correct referred objects. For the examples in Fig.~\ref{fig:mul}, the high-level features is capable of roughly identifying the object of interest and gradually refine the response with fine boundary from the second column to fourth column. 
Fig.~\ref{fig:words_vis} presents visualization results when our network responses to diverse referring expressions that are not included in the dataset. The purpose of showing these examples is to verify the generalization ability of our network. The visualization results show that our network can effectively handle different concepts appearing in the expressions and accurately identify the referred objects under different semantics.

Attention models have been widely used and proven important for image captioning~\cite{xu2015show,lu2017knowing} to relate a part of active region to each generated word in the caption. These attention-based methods follow a similar encoder-decoder framework which encodes spatial image features and the hidden state of the word feature to produce the visual context, then decodes the context to predict the next word instead of a segmentation mask in referring image segmentation task. Their methods also have the ability to visualize the related regions of the image through word and context information to illustrate which image region the model is focusing on. In order to compare with attention model of image captioning for visual and linguistic features, we adopt the soft attention model~\cite{xu2015show} in our encoder network and show visualization results in Fig.~\ref{fig:comp_cap}. It can be observed that the highlighted regions by the attention model of image captioning shown in Fig.~\ref{fig:comp_cap}~(b) are more scattered and do not present a clear boundary of the referred object-of-interest since their method loses spatial information when producing the visual context. These attention maps might be sufficient for the image captioning task, but do not provide precise spatial information needed in the referring image segmentation task. As shown in Fig.~\ref{fig:comp_cap}~(c), our E-ConvLSTM precisely identifies the referred object and suppresses the background regions, since our encoder network can keep the spatial information of the features and embed the word attention in the multimodal interaction by ConvLSTM  to adaptively localize the referred objects.

We further provide qualitative examples to illustrate the effect of the spatial features. Fig.~\ref{fig:wo_sp} presents visualization results generated by our network with and without the spatial features. As shown in Fig.~\ref{fig:wo_sp}~(b), the network without the spatial feature can be confused by the objects with similar appearance, while they can be clearly discriminated by the network with spatial features. For example, the referred hotdog is located at the bottom of the another as shown in the first row and the desired donut is at the relative location compared to other donuts in the second row. 

\begin{figure}[htb]
  \centering
{\setlength{\tabcolsep}{0.5pt}
\scalebox{0.9}{
  \begin{tabular}{cccccc}

   \multicolumn{3}{c}{Referring expression: \textit{``a sparrow is sitting along with two others''}} \\
    \includegraphics[width=1.2in]{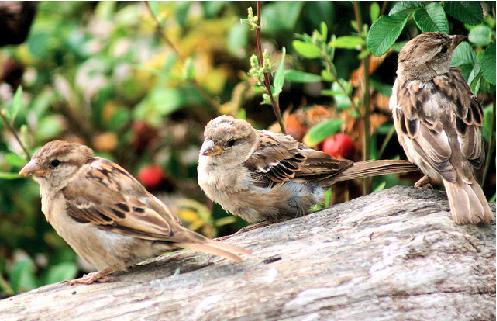}&
    \includegraphics[width=1.2in]{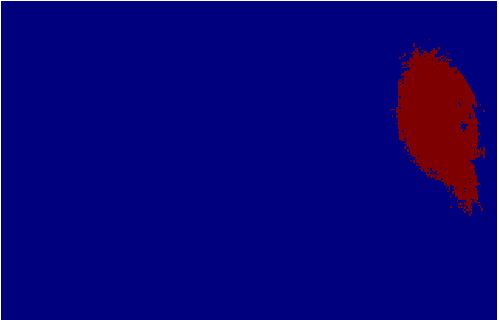}&
      \includegraphics[width=1.2in]{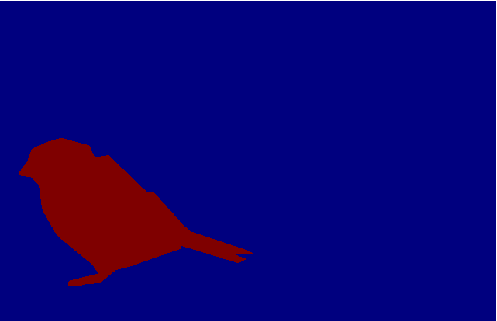}\\

    \multicolumn{3}{c}{Referring expression: \textit{``man player on right''}} \\
    \includegraphics[width=1.2in]{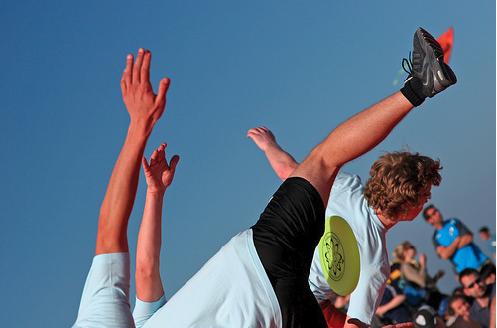}&
    \includegraphics[width=1.2in]{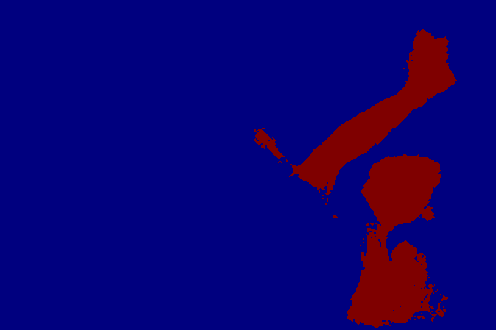}&
      \includegraphics[width=1.2in]{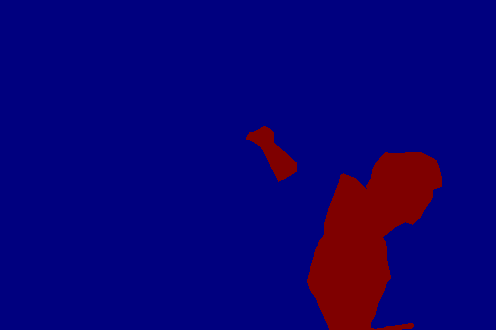}\\
      
    \multicolumn{3}{c}{Referring expression: \textit{``the bear hiding behind pole''}} \\
    \includegraphics[width=1.2in]{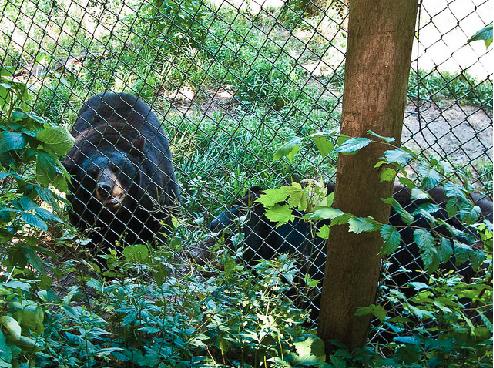}&
    \includegraphics[width=1.2in]{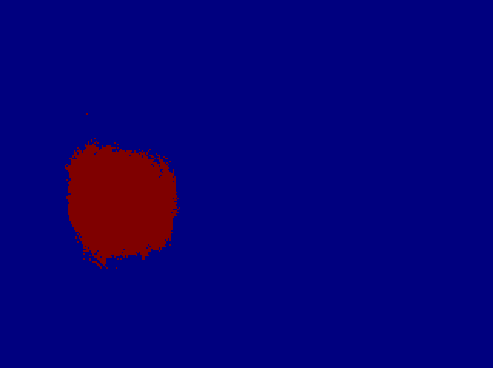}&
      \includegraphics[width=1.2in]{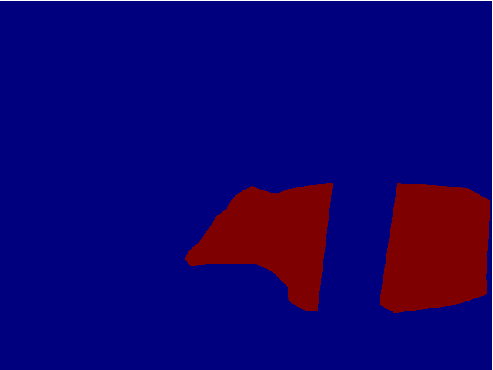}\\
      
               \multicolumn{3}{c}{Referring expression: \textit{``left brocoli''}} \\
    \includegraphics[width=1.2in]{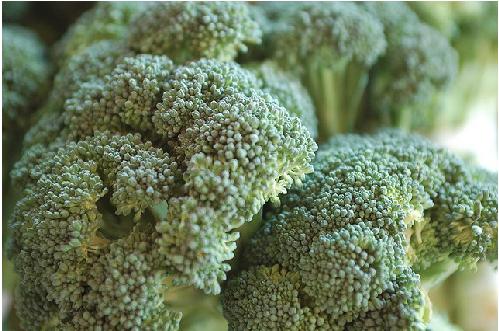}&
    \includegraphics[width=1.2in]{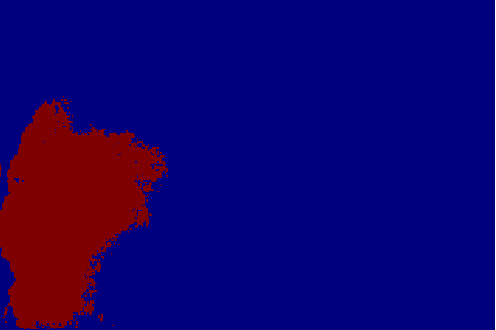}&
      \includegraphics[width=1.2in]{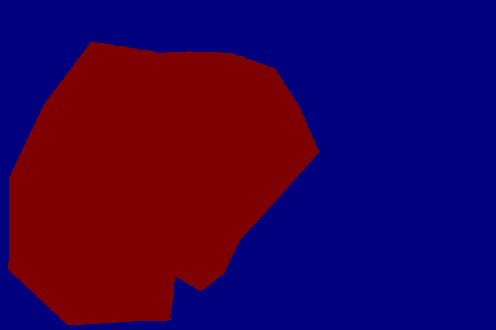}\\

      (a) & (b) & (c)\\
    
  \end{tabular}
}
}

\caption{Some examples of failure cases of our network. (a) original image; (b) predicted segmentation mask; (c) ground truth segmentation mask.}
\label{fig:failure}
\end{figure}

Some examples of failure cases are presented in Fig.~\ref{fig:failure}. In the first example, our network may have failed to identify the exact object due to the language ambiguity where the predicted segmentation mask points to a sparrow different from the ground truth. 
The second example shows the case where our method correctly understands the referring expression and recognize the attributes of a person, e.g., head, hand and leg, but incorrectly segments the leg on the right which actually belongs to another person on the left. In the third example, our segmentation result is negatively affected by the heavy occlusion to find the bear. 
In the last example, the failure is caused by some internal gaps within the broccoli that misleads our network to consider they are apart and only segment the left part of the referred broccoli.
We are going to explore fine-grained recognition technique to detect these subtle details in the future.

\section{Conclusion}\label{sec:conclude}
We present a novel dual convolutional LSTM network for the task of referring image segmentation. The E-ConvLSTM encodes the multimodal feature along word sequence to localize the referred objects and D-ConvLSTM decodes the 
encoded multimodal information at multiple levels for mask refinement. Extensive experiments on four datasets shows consistent improvement achieved by the proposed network.



\bibliographystyle{IEEEtran}
\bibliography{J_ref}


\end{document}